\documentclass[letterpaper]{article} 
\usepackage{aaai23}  
\usepackage{times}  
\usepackage{helvet}  
\usepackage{courier}  
\usepackage[hyphens]{url}  
\usepackage{graphicx} 
\usepackage{natbib}  
\usepackage{caption} 
\usepackage{algorithm}
\usepackage{algorithmic}

\usepackage{amsmath,amsfonts,bm}
\usepackage{dsfont}
\usepackage{booktabs}       
\usepackage{subfigure}
\usepackage{multirow}
\usepackage{adjustbox}

\usepackage{array}
\newcolumntype{R}[2]{%
    >{\adjustbox{angle=#1,lap=\width-(#2)}\bgroup}%
    l%
    <{\egroup}%
}
\newcommand*\rot{\multicolumn{1}{R{35}{1em}}}

\newcommand{\ie}{\textit{i}.\textit{e}., }
\newcommand{\eg}{\textit{e}.\textit{g}., }
\newcommand{\viz}{\textit{viz}., }
\newcommand{\pz}{\phantom{0}}
\newcommand{\stdv}[1]{\scriptsize~$\pm$~#1}

\usepackage{pifont}  
\newcommand{\cmark}{\ding{51}}
\newcommand{\xmark}{\ding{55}}

\DeclareMathOperator*{\argmax}{arg\,max}

\usepackage{newfloat}
\usepackage{listings}
\DeclareCaptionStyle{ruled}{labelfont=normalfont,labelsep=colon,strut=off} 
\floatstyle{ruled}
\newfloat{listing}{tb}{lst}{}
\floatname{listing}{Listing}
\title{Confidence-aware Training of Smoothed Classifiers for Certified Robustness}
\author{
    Jongheon Jeong\equalcontrib, Seojin Kim\equalcontrib, Jinwoo Shin
}
\affiliations{
    Korea Advanced Institute of Science and Technology (KAIST)\\

    Daejeon, 34141 South Korea\\
    \{jongheonj, osikjs, jinwoos\}@kaist.ac.kr
}

\begin{document}

\maketitle

\begin{abstract}
Any classifier can be ``smoothed out'' under Gaussian noise to build a new classifier that is provably robust to $\ell_2$-adversarial perturbations, \viz by averaging its predictions over the noise via \emph{randomized smoothing}. Under the \emph{smoothed classifiers}, the fundamental trade-off between accuracy and (adversarial) robustness has been well evidenced in the literature: \ie increasing the robustness of a classifier for an input can be at the expense of decreased accuracy for some other inputs. In this paper, we propose a simple training method leveraging this trade-off to obtain robust smoothed classifiers, in particular, through a \emph{sample-wise} control of robustness over the training samples. We make this control feasible by using ``accuracy under Gaussian noise'' as an easy-to-compute proxy of adversarial robustness for an input. Specifically, we differentiate the training objective depending on this proxy to filter out samples that are unlikely to benefit from the worst-case (adversarial) objective. Our experiments show that the proposed method, despite its simplicity, consistently exhibits improved certified robustness upon state-of-the-art training methods. Somewhat surprisingly, we find these improvements persist even for other notions of robustness, \eg to various types of common corruptions. Code is available at \url{https://github.com/alinlab/smoothing-catrs}.
\end{abstract}

\section{Introduction}

Despite these tremendous advances in \emph{deep neural networks} for a variety of computer vision tasks towards artificial intelligence, the broad existence of \emph{adversarial examples} \cite{szegedy2013intriguing} is still a significant aspect that reveals the gap between machine learning systems and humans: for a given input $x$ (\eg an image) to a classifier $f$, say a neural network, $f$ often permits a perturbation $\delta$ that completely flips the prediction $f(x+\delta)$, while $\delta$ is too small to change the semantic in $x$. In response to this vulnerability, there have been tremendous efforts in building \emph{robust} neural network based classifiers against adversarial examples, either in forms of \emph{empirical defenses} \cite{pmlr-v80-athalye18a,carlini2019evaluating,tramer2020adaptive}, which are largely based on \emph{adversarial training} \cite{madry2018towards,pmlr-v97-zhang19p,Wang2020Improving,pmlr-v119-zhang20z,wu2020awp}, or \emph{certified defenses} \cite{pmlr-v80-wong18a,xiao2018training,pmlr-v97-cohen19c,zhang2020towards}, depending on whether the robustness claim can be theoretically guaranteed or not.

\emph{Randomized smoothing} \cite{lecuyer2019certified,pmlr-v97-cohen19c}, our focus in this paper, is currently a prominent approach in the context of certified defense, thanks to its scalability to arbitrary neural network architectures while previous methods have been mostly limited in network sizes or require strong assumptions, \eg Lipschitz constraint, on their architectures: specifically, for a given classifier $f$, it constructs a new classifier $\hat{f}$, where $\hat{f}(x)$ is defined to be the class that $f(x+\delta)$ outputs most likely over $\delta \sim \mathcal{N}(0, \sigma^2 I)$, \ie the Gaussian noise. Then, it is shown by \citet{lecuyer2019certified} that $\hat{f}$ is certifiably robust in $\ell_2$-norm, and \citet{pmlr-v97-cohen19c} further tightened the $\ell_2$-robustness guarantee which is currently considered as the state-of-the-art in certified defense. 

However, even with recent methods for adversarial defense, including randomized smoothing, the \emph{trade-off} between robustness and accuracy \cite{tsipras2018robustness,pmlr-v97-zhang19p} has been well evidenced, \ie increasing the robustness for a specific input can be at the expense of decreased accuracy for other inputs. For instance, with the current best practices, \citet{salman2020adversarially} reports that the accuracy of ResNet-50 on ImageNet degrades, \eg 75.8\% $\rightarrow$ 63.9\%, by an $\ell_\infty$-adversarial training, \ie optimizing the classifier to ensure robustness at all the given training samples around an $\ell_\infty$-ball of size $\tfrac{4}{255}$. In addition, \citet{pmlr-v97-zhang19p} has shown that the (empirical) robustness of a classifier can be further boosted in training by paying more expense in accuracy. A similar trend can be also observed with certified defenses, \eg randomized smoothing, as the clean accuracy of smoothed classifiers are usually less than those one can obtain from the standard training on the same architecture \cite{pmlr-v97-cohen19c}. 

\begin{figure}[t]
\centering
    \hspace*{\fill}
	\subfigure[Bottom-$K$ loss]
	{
	    \includegraphics[height=1.2in]{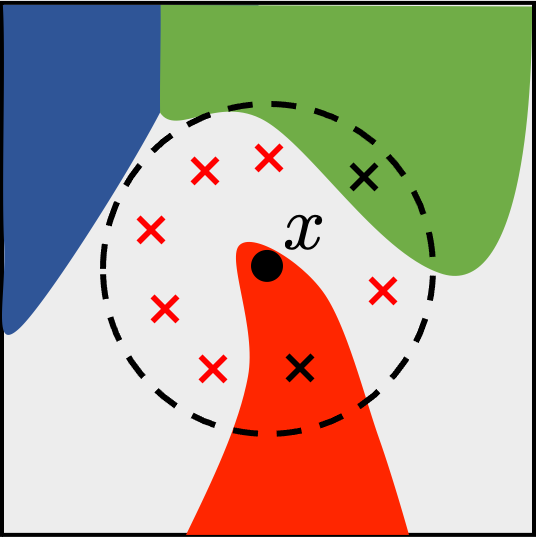}
		\label{fig:low_conf}
	}
	\hspace*{\fill}
	\subfigure[Worst-case loss]
	{
	    \includegraphics[height=1.2in]{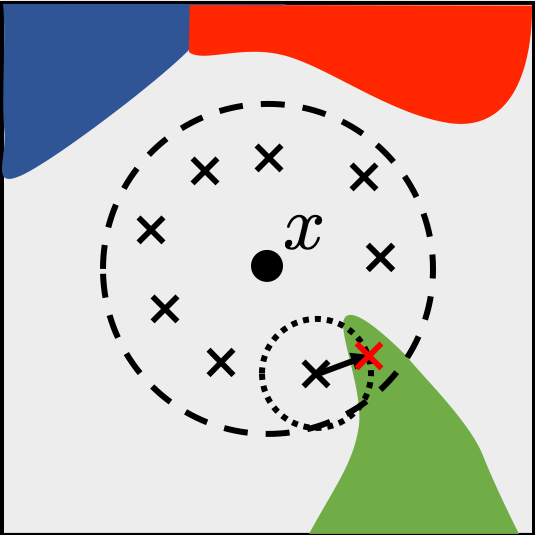}
		\label{fig:high_conf}
	}
	\hspace*{\fill}
	\caption{Illustration of the two proposed losses, \ie the (a) \emph{bottom-$K$} and (b) \emph{worst-case} losses. Each $\times$ represents Gaussian noise around $x$. {We aim to 
	minimize the cross-entropy loss only for $\times$'s marked as red for each case.}}
	\label{fig:overview}
\end{figure}

\paragraph{Contribution.} In this paper, we develop a novel training method for randomized smoothing, coined \emph{Confidence-Aware Training for Randomized Smoothing} (CAT-RS), which {incorporates a \emph{sample-wise} control of target robustness on-the-fly motivated by the accuracy-robustness trade-off in smoothed classifiers.}
Intuitively, a natural approach one can consider in response to the trade-off in robust training is to appropriately lower the robustness requirement for ``hard-to-classify'' samples while maintaining those for the remaining (``easier'') samples: here, the challenges are (a) which samples should we choose as either {``hard-to-classify'' (or ``easier'')} for the control in training, and (b) how to control their target robustness. For both (a) and (b), the major difficultly stems from that evaluating adversarial robustness for a given sample is computationally hard in practice. 

To implement this idea, we focus on a peculiar correspondence from \emph{prediction confidence} to adversarial robustness that smoothed classifiers offer: due to its local-Lipschitzness \cite{nips_salman19}, achieving a high confidence at $x$ from a smoothed classifier also implies a high (certified) robustness at $x$. 
Inspired by this, we propose to use the sample-wise confidence of smoothed classifiers as an efficient proxy of the certified robustness, 
and defines two new losses, namely the \emph{bottom-$K$} and \emph{worst-case} Gaussian training, each of those targets different levels of confidence so that the overall training can prevent low-confidence samples from being enforced to increase their robustness. 

We verify the effectiveness of our proposed method through an extensive comparison with existing robust training methods for smoothed classifiers, including the state-of-the-arts, on a wide range of benchmarks on MNIST, Fashion-MNIST, CIFAR-10/100, and ImageNet. Our experimental results constantly show that the proposed method can significantly improve the previous state-of-the-art results on certified robustness achievable from a given neural network architecture, by (a) maximizing the robust radii of high-confidence samples while (b) reducing the risk of deteriorating the accuracy at low-confidence samples. More intriguingly, we also observe that such a training scheme also helps smoothed classifiers to generalize beyond adversarial robustness, as evidenced by significant improvements in robustness against common corruptions compared to other robust training methods. Our extensive ablation study further confirms that each of both proposed components has an individual effect on improving certified robustness, and can effectively control the accuracy-robustness trade-off with the hyperparameter between the two proposed losses.

\paragraph{Related work.} There have been continual attempts to provide a certificate on robustness of deep neural networks against adversarial attacks \cite{gehr2018ai2,pmlr-v80-wong18a,pmlr-v80-mirman18b,xiao2018training,gowal2019scalable,zhang2020towards}, and correspondingly to further improve the robustness with respect to those certification protocols \cite{pmlr-v89-croce19a,Croce2020Provable,Balunovic2020Adversarial}.\footnote{A more extensive survey on certified robustness can be found in \citet{li2020sokcertified}.} \emph{Randomized smoothing} \cite{pmlr-v97-cohen19c} has attracted a particular attention among them, due to its scalability to large datasets and its flexibility to various applications \cite{pmlr-v119-rosenfeld20b,salman2020denoised,wang2021graph,pmlr-v139-fischer21a,wu2022crop} or other threat models \cite{li2021tss,pmlr-v119-yang20c,lee2019tight,jia2020certified,zhang2020blackbox,salman2021certified}. 

This work aims to improve adversarial robustness of randomized smoothing, along a line of research on designing training schemes specialized for smoothed classifiers \cite{nips_salman19,Zhai2020MACER,jeong2020consistency,jeong2021smoothmix}. Specifically, we focus on the relationship between confidence and robustness of smoothed classifiers, a property rarely investigated previously but few \cite{kumar2020certifying,jeong2021smoothmix}. We leverage the property to overcome challenges in estimating sample-wise robustness, and to develop a data-dependent adversarial training which has been also challenging even for empirical robustness \cite{Wang2020Improving,zhang2021geometryaware}.

\section{Preliminaries}

\paragraph{Adversarial robustness.} Consider a labeled dataset $\mathcal{D}=\{(x_i, y_i)\}^n_{i=1}$ sampled from $P$, where $x \in \mathbb{R}^d$ and $y \in \mathcal{Y}:=\{1, \cdots, K\}$, and let $f: \mathbb{R}^d\rightarrow \mathcal{Y}$ be a classifier. Given that $f$ is discrete, one can consider a differentiable $F: \mathbb{R}^{d}\rightarrow \Delta^{K-1}$ to allow a gradient-based optimization assuming $f(x):=\argmax_{k\in\mathcal{Y}}F_k (x)$, where $\Delta^{K-1}$ is probability simplex in $\mathbb{R}^K$. The standard framework of \emph{empirical risk minimization} to optimize $f$ assumes that the samples in $\mathcal{D}$ are \textit{i.i.d.}\ from $P$ and expect $f$ to perform well given that the future samples also follow the \textit{i.i.d.}\ assumption.

However, in the context of \emph{adversarial robustness} (and for other notions of robustness as well), the \textit{i.i.d.}\ assumption on the future samples does not hold anymore: instead, it assumes that the samples can be \emph{arbitrarily} perturbed up to a certain restriction, \eg a bounded $\ell_2$-ball, and focuses on the \emph{worst-case} performance over the perturbed samples. One way to quantify this is the \emph{average minimum-distance} of adversarial perturbation \cite{moosavi2016deepfool,carlini2019evaluating}:
\begin{equation}
\label{eq:avg_min_dist}
    R(f; P) := \mathbb{E}_{(x, y)\sim P}\left[\min_{f (x')\ne y} ||x' - x||_2\right].
\end{equation}

\paragraph{Randomized smoothing.} The essential challenge in achieving adversarial robustness in neural networks, however, stems from that directly evaluating \eqref{eq:avg_min_dist} (and further optimizing it) is usually computationally infeasible, \eg under the standard practice that $F$ is modeled by a complex, high-dimensional neural network. \emph{Randomized smoothing} \cite{lecuyer2019certified,pmlr-v97-cohen19c} bypasses this difficulty by constructing a new classifier $\hat{f}$ from $f$ instead of letting $f$ to directly model the robustness: specifically, it transforms the base classifier $f$ with a certain \emph{smoothing measure}, where in this paper we focus on the case of Gaussian distributions $\mathcal{N}(0, \sigma^2 I)$: 
\begin{equation}
\label{eq:smoothing}
    \hat{f}(x) := \argmax_{c\in \mathcal{Y}} \mathbb P_{\delta\sim\mathcal N(0,\sigma^2I)}\left(f(x+\delta)=c\right).
\end{equation}
Then, the robustness of $\hat{f}$ at $(x, y)$, namely $R(\hat{f}; x, y)$, can be lower-bounded in terms of the \emph{certified radius} $\underline{R}(\hat{f}, x, y)$, \eg \citet{pmlr-v97-cohen19c} showed that the following bound holds which is tight for $\ell_2$-adversary:
\begin{align}
    \label{eq:cr}
    R(\hat{f}; x, y) &\ge \sigma \cdot \Phi^{-1}(p_f(x, y)) =: \underline{R}(\hat{f}, x, y) \\
    \text{where}\quad p_f(x, y) &:= {\mathbb{P}_{\delta}(f(x+\delta)=y)},
    \label{eq:p_f}
\end{align}
provided that $\hat{f}(x) = y$, otherwise $R(\hat{f}; x, y) := 0$.\footnote{$\Phi$ denotes the cumulative distribution function of $\mathcal{N}(0,1^2)$.} Here, we remark that the formula for certified radius \eqref{eq:cr} is essentially a function of $p_f$ \eqref{eq:p_f}, which represents the \emph{prediction confidence} of $\hat{f}$ at $x$, or equivalently, the \emph{accuracy} of $f(x+\delta)$ over $\delta\sim \mathcal{N}(0, \sigma^2 I)$. In other words, unlike standard neural networks, smoothed classifiers can guarantee a correspondence from prediction confidence to adversarial robustness - which is the key motivation of our method. 

\section{Confidence-aware Randomized Smoothing}
\label{s:method}

We aim to develop a new training method to maximize the certified robustness of a smoothed classifier $\hat{f}$, considering the trade-off relationship between robustness and accuracy \cite{pmlr-v97-zhang19p}: even though randomized smoothing can be applied for any classifier $f$, the actual robustness of $\hat{f}$ depends on how much $f$ classifies well under presence of Gaussian noise, \ie by $p_f(x, y)$ defined in \eqref{eq:p_f}. A simple way to train $f$ for a robust $\hat{f}$, therefore, is to minimize the cross-entropy loss (denoted by $\mathbb{CE}$ below) with Gaussian augmentation as in \citet{pmlr-v97-cohen19c}: 
\begin{equation}
\label{eq:gaussian_training}
    \min_{F}\ \mathbb{E}_{\substack{(x, y)\sim P \\ \delta\sim\mathcal{N}(0, \sigma^2 I)}} \left[\mathbb{CE}(F(x+\delta), y)\right].
\end{equation}

In this paper, we extend this basic form of training to incorporate a \emph{confidence-aware} strategy to decide which noise samples $\delta_i \sim\mathcal{N}(0, \sigma^2 I)$ should be used sample-wise for training $f$. Ideally, one may wish to obtain a classifier $f$ that achieves $p_f(x, y)\approx 1$ for every $(x, y) \sim P$ to maximize its certified robustness. In practice, however, such a case is highly unlikely, and there usually exists a sample $x$ that $p_f(x, y)$ should be quite lower than 1 to maintain the discriminativity with other samples: in other words, these samples can be actually ``beneficial'' to be misclassified at some (hard) Gaussian noises, otherwise the classifier has to memorize the noises to correctly classify them. On the other hand, for the samples which can indeed achieve $p_f(x, y)\approx 1$, the current Gaussian training \eqref{eq:gaussian_training} may not be able to provide enough samples of $\delta_i$ for $x$ throughout the training, as $p_f(x, y)\approx 1$ implies that $f(x+\delta)$ must be correctly classified ``almost surely'' for $\delta_i \sim\mathcal{N}(0, \sigma^2 I)$. 

In these respects, we propose two different variants of Gaussian training \eqref{eq:gaussian_training} that address each of the possible cases, \ie whether (a) $p_f(x, y) < 1$ or (b) $p_f(x, y)\approx 1$, namely with (a) \emph{bottom-$K$} and (b) \emph{worst-case} Gaussian training, respectively. During training, the method first estimates $p_f(x, y)$ for each sample by computing their accuracy over $M$ random samples of $\delta\sim\mathcal{N}(0, \sigma^2 I)$, and applies different forms of loss depending on the value. In the following two sections, Section~\ref{ss:bottom_K} and \ref{ss:worst_case}, we provide the details on each loss, and Section~\ref{ss:overall} describes how to combine the two losses and defines the overall training scheme.

\subsection{Bottom-$K$ Loss for Low-confidence Samples}
\label{ss:bottom_K}

Consider a base classifier $f$ and a training sample $(x, y) \in \mathcal{D}$, and suppose that $p_f(x, y) \ll 1$, \eg $\hat{f}$ has a low-confidence at $x$. Figure~\ref{fig:low_conf} visualizes this scenario: in this case, by definition of $p_f(x, y)$ in \eqref{eq:p_f}, $f(x+\delta)$ would be correctly classified to $y$ only with probability $p$ over $\delta\sim \mathcal{N}(0, \sigma^2 I)$, and this implies either (a) $x + \delta$ has not yet been adequately exposed to $f$ during the training, or (b) $x + \delta$ may be indeed hard to be correctly classified for some $\delta$, so that minimizing the loss at these noises could harm the generalization of $\hat{f}$. The design goal of our proposed \emph{bottom-$K$ Gaussian loss} is to modify the standard Gaussian training \eqref{eq:gaussian_training} to reduce the optimization burden from (b) while minimally retaining its ability to cover enough noise samples during training for (a).

We first assume $M$ random \textit{i.i.d.}~samples of $\delta$, say $\delta_1, \delta_2,  \cdots,  \delta_M \sim \mathcal{N}(0, \sigma^2 I)$. One can notice that the random variables $\mathds{1}[f(x+\delta_i)=y]$'s are also \textit{i.i.d.} each, which follows the Bernoulli distribution of probability $p_f(x, y)$. This means that, if the current $p_f(x, y)$ is the value one attempts to keep instead of further increasing it, the number of ``correct'' noise samples, namely $\sum_i \mathds{1}[f(x+\delta_i)=y]$, would follow the \emph{binomial distribution} $K\sim \mathrm{Bin}(M, p)$ - this motivates us to consider the following loss that only minimizes the \emph{$K$-smallest} cross-entropy losses out of from $M$ Gaussian samples around $x$: 
\begin{equation}
\label{eq:bottom_k}
    L^{\tt low} := 
    \frac{1}{M} \sum_{i=1}^{K} \mathbb{CE}(F(x+\delta_{\pi(i)}), y), 
\end{equation}
where $K \sim \mathrm{Bin}(M, p_f(x, y))$. Here, $\pi(i)$ denotes the index with the $i$-th smallest loss value in the $M$ samples. 

Yet, the loss defined in \eqref{eq:bottom_k} may not handle the \emph{cold-start} problem on $p_f(x, y)$, \eg at the early stage of the training where $x+\delta$ has not been adequately exposed to $f$, so that it is uncertain whether the current $p_f(x, y)$ is optimal: in this case, $L^{\tt low}$ can be minimized with an under-estimated $p_f \approx 0$, potentially with samples those never optimize the cross-entropy losses during training. Nevertheless, we found that a simple workaround of \emph{clamping} $K$ can effectively handle the issue, \ie by using $K^+ \leftarrow \max(K, 1)$ instead of $K$: in other words, we always allow the ``easiest'' noise among the $M$ samples to be fed into $f$ throughout the training.

\subsection{Worst-case Loss for High-confidence Samples}
\label{ss:worst_case}

Next, we focus on the case when $p_f(x, y) \approx 1$, \ie $\hat{f}$ has a high confidence at $x$,  as illustrated in Figure~\ref{fig:high_conf}. In contrast to the previous scenario in Section~\ref{ss:bottom_K} (and Figure~\ref{fig:low_conf}), now the major drawback of Gaussian training \eqref{eq:gaussian_training} does not come from the \emph{abundance} of hard noises in training, but from the \emph{rareness} of such noises: considering that one can only present a limited number of noise samples to $f$ throughout its training, na\"ively minimizing \eqref{eq:gaussian_training} may not cover some ``potentially hard'' noise samples, and this would result in a significant harm in the final certified radius of the smoothed classifier $\hat{f}$. The purpose of \emph{worst-case} Gaussian training is to overcome this lack of samples via an \emph{adversarial} search around each of the noise samples.

Specifically, for given $M$ samples of Gaussian noise $\delta_i$ as considered in \eqref{eq:bottom_k}, namely $\delta_1, \delta_2, \cdots, \delta_M \sim\mathcal{N}(0, \sigma^2 I)$, we propose to modify \eqref{eq:gaussian_training} to find the \emph{worst-case} noise $\delta^*$ (a) around an $\ell_2$-ball for each noise as well as (b) among the $M$ samples, and minimize the loss at $\delta^*$ instead of the average-case loss. To find such worst-case noise, our proposed loss optimizes a given $\delta_i$ to maximize the \emph{consistency} of its prediction from a certain label assignment $\hat{y}\in \Delta^{K-1}$ per $x$:
\begin{equation}
\label{eq:worst}
    L^{\tt high} := \max_i \max_{\|\delta^*_i - \delta_i \|_2 \le \varepsilon}  \mathrm{KL}(F(x+\delta^*_i), \hat{y}), 
\end{equation}
where $\mathrm{KL}(\cdot, \cdot)$ denotes the Kullback-Leibler divergence. This objective is motivated by \cite{jeong2020consistency} that the consistency of prediction across different Gaussian noise controls the trade-off between accuracy and robustness of smoothed classifiers. Notice from \eqref{eq:worst} that the objective is equivalent to the cross-entropy loss if $\hat{y}$ is assigned as (hard-labeled) $y$, while we observe having a soft-labeled $\hat{y}$ is beneficial in practice: its log-probability, where the consistency targets, can now be bounded so $F(x+\delta^*_i)$'s can also minimize their variance in the logit space. 

There can be various ways to assign $\hat{y}$ for a given $x$. One reasonable strategy, which we use in this paper by default, is to assign $\hat{y}$ by the \emph{smoothed prediction} of another classifier $\bar{f}$, pre-trained on $\mathcal{D}$ via Gaussian training \eqref{eq:gaussian_training} with some $\sigma_0$. This approach is (a) easy to compute, and (b) naturally reflects sample-wise difficulties under Gaussian noise, while (c) maintaining the label information from $y$. 
Nevertheless, we also confirm in Appendix~\ref{ap:abla_loss} that $L^{\tt high}$ is still effective even when $\hat{y}$ is defined in a simpler way, namely by the average of $F(x+\delta_i)$'s without the Gaussian pre-training.

In practice, we use the \emph{projected gradient descent} (PGD) \cite{madry2018towards} to solve the inner maximization in \eqref{eq:worst}: namely, we perform a $T$-step gradient ascent from each $\delta_i$ with step size ${2\cdot\varepsilon} / {T}$ while projecting the perturbations to be in the $\ell_2$-ball of size $\varepsilon$. This procedure would find a noise $\delta^*$ that maximizes the loss around $x$, while maintaining the Gaussian-like noise appearance due to the projected search in a small $\varepsilon$-ball. In order to further make sure that the Gaussian likelihood of $\delta^*$ is maintained from the original $\delta$, we additionally apply a simple trick of \emph{normalizing} the mean and standard deviation of $\delta^*$ to follow those of $\delta$.

\paragraph{Comparison to SmoothAdv.} The idea of incorporating an adversarial search for the robustness of smoothed classifiers has been also considered in previous works \cite{nips_salman19,jeong2021smoothmix}: \eg \citet{nips_salman19} have proposed \emph{SmoothAdv} that applies adversarial training \cite{madry2018towards} to a ``soft'' approximation of $\hat{f}$ given $f$ and $M$ noise samples:
\begin{equation}
\label{eq:smoothadv}
    x^* = \argmax_{||x' - x||_2 \le \epsilon}\left(-\log \left(\frac{1}{M} \sum_i F_y(x' +\delta_i)\right)\right).
\end{equation}
Our method is different from the previous approaches in which part of the inputs is adversarially optimized: \ie we directly optimize the noise samples $\delta_i$'s instead of $x$, with no need to assume a soft relaxation of $\hat{f}$. This is due to our unique motivation of finding the worst-case Gaussian noise, and our experimental results in Section~\ref{s:experiments} further support the effectiveness of this approach.  

\subsection{Overall Training Scheme}
\label{ss:overall}

Given the two losses $L^{\tt low}$ and $L^{\tt high}$ defined in Section~\ref{ss:bottom_K} and \ref{ss:worst_case}, respectively, we now define the full objective of our proposed \emph{Confidence-Aware Training for Randomized Smoothing} (CAT-RS). Overall, in order to differentiate how to combine the two losses per sample basis, we use the smoothed confidence $p_f(x, y)$ \eqref{eq:p_f} as the guiding proxy: specifically, we aim to apply the worst-case loss of $L^{\tt high}$ only for the samples where $p_f(x, y)$ is already high enough. In practice, however, one does not have a direct access to the value of $p_f(x, y)$ during training, and we estimate this with the $M$ noise samples\footnote{We use $M=4$ for our method unless otherwise noted.} as done for $L^{\tt low}$ and $L^{\tt high}$, \ie by $\hat{p}_f(x, y) := \frac{1}{M} \sum_{i=1}^{M} \mathds{1}[f(x+\delta_i)=y]$. Then, we consider a simple and intuitive masking condition of ``$K=M$'' to activate $L^{\tt high}$, where $K\sim \mathrm{Bin}(M, \hat{p}_f(x, y))$ is the random variable defined in \eqref{eq:bottom_k} for $L^{\tt low}$. The final loss becomes:
\begin{equation}
\label{eq:overall}
    L^{\mathtt{CAT}\text{-}\mathtt{RS}} := L^{\tt low} + \lambda \cdot \mathds{1}[K=M] \cdot L^{\tt high},
\end{equation}
where $\mathds{1}[\cdot]$ is the indicator random variable, and $\lambda > 0$. In other words, the training minimizes $L^{\tt high}$ only when $L^{\tt low}$ \eqref{eq:bottom_k} minimizes the ``full'' cross-entropy losses for all the $M$ noise samples given around $(x, y)$. 
The hyperparameter $\lambda$ in \eqref{eq:overall} controls the trade-off between accuracy and robustness \cite{pmlr-v97-zhang19p} of CAT-RS: given that $L^{\tt high}$ targets samples that achieves high confidence (\ie they are already robust), having larger weights on $L^{\tt high}$ results in higher certified robustness at large radii.
In terms of computational complexity, the proposed CAT-RS takes a similar training cost with recent methods those also perform adversarial searches with smoothed classifiers, \eg SmoothAdv \cite{nips_salman19} and SmoothMix \cite{jeong2021smoothmix}.\footnote{A comparison of actual training costs is given in Appendix~\ref{ap:training_cost}.} The complete procedure of computing our proposed CAT-RS loss can be found in Algorithm~\ref{alg:training} of Appendix~\ref{ap:alg}.

\begin{table*}[t]
\centering
\small
\begin{adjustbox}{width=0.8\linewidth}
\begin{tabular}{clc|ccccccccccc}
    \toprule
    
    $\sigma$ &  Methods & ACR & 0.00 & 0.25 & 0.50 & 0.75 & 1.00 & 1.25 & 1.50 & 1.75 & 2.00 & 2.25 & 2.50 \\ 
    \midrule
    \multirow{7.5}{*}{0.25}& Gaussian  & 0.424 & 76.6 & 61.2 & {42.2} & {25.1} & 0.0 & 0.0 & 0.0 & 0.0& 0.0 & 0.0& 0.0\\
    & Stability & 0.420 & {73.0} & {58.9} & 42.9 & 26.8 & 0.0 & 0.0 & 0.0 & 0.0 & 0.0 & 0.0 & 0.0 \\ 
    & SmoothAdv  & 0.544 & 73.4 & 65.6 & 57.0 & {47.5} & 0.0 & 0.0 & 0.0 & 0.0 & 0.0 & 0.0 & 0.0  \\
    & MACER  & 0.531 & \underline{79.5} & 69.0 & 55.8 & 40.6 & 0.0 & 0.0 & 0.0 & 0.0 & 0.0 & 0.0 & 0.0 \\  
    & Consistency & 0.552 & 75.8 & 67.6 & 58.1 & 46.7 & 0.0 & 0.0 & 0.0 & 0.0 & 0.0 & 0.0 & 0.0 \\
    & SmoothMix  & 0.553 & 77.1 & 67.9 & 57.9 & 46.7 & 0.0 & 0.0 & 0.0 & 0.0 & 0.0 & 0.0 & 0.0 \\
    \cmidrule(l){2-2} \cmidrule(l){3-3} \cmidrule(l){4-14}
    
    & \textbf{CAT-RS (Ours)} & \underline{\textbf{0.562}} & 76.3 & \underline{\textbf{68.1}} & \underline{\textbf{58.8}} & \underline{\textbf{48.2}} & 0.0 & 0.0 & 0.0 & 0.0 & 0.0 & 0.0 & 0.0 \\

    \midrule
    \multirow{7.5}{*}{0.50}& Gaussian & 0.525 & \underline{65.7} & 54.9 & 42.8 & {32.5} & {22.0} & {14.1} & {8.3} & {3.9} & 0.0 & 0.0 & 0.0\\
    & Stability & 0.531 & 62.1 & {52.6} & {42.7} & 33.3 & 23.8 & 16.1 & 9.8 & 4.7 & 0.0 & 0.0 & 0.0\\
    & SmoothAdv & 0.684 & 65.3 & \underline{57.8} & 49.9 & 41.7 & 33.7 & 26.0 & 19.5 & 12.9 & 0.0 & 0.0  & 0.0\\
    & MACER & 0.691  & 64.2 & {57.5} & 49.9 & 42.3 & 34.8 & 27.6 & 20.2 & 12.6 & 0.0 & 0.0 & 0.0\\
    & Consistency & 0.720 & {64.3} & {57.5} & \underline{50.6} & 43.2 & 36.2 & 29.5 & 22.8 & 16.1 & 0.0 & 0.0 & 0.0 \\ 
    & SmoothMix  & 0.737 & {61.8} & 55.9 & 49.5 & 43.3 & 37.2 & 31.7 & 25.7 & 19.8 & 0.0 & 0.0 & 0.0 \\
    \cmidrule(l){2-2} \cmidrule(l){3-3} \cmidrule(l){4-14}
    & \textbf{CAT-RS (Ours)} & \underline{\textbf{0.757}} & 62.3 & \textbf{56.8} & \textbf{50.5} & \underline{\textbf{44.6}} & \underline{\textbf{38.5}} & \underline{\textbf{32.7}} & \underline{\textbf{27.1}} & \underline{\textbf{20.6}} & 0.0 & 0.0 & 0.0\\

    \midrule
    \multirow{7.5}{*}{1.00}& Gaussian & 0.511 & \underline{47.1} & 40.9 & 33.8 & 27.7 & {22.1} & {17.2} & {13.3} & {9.7} & {6.6} & {4.3} & {2.7} \\
    & Stability  & 0.514 & 43.0 & {37.8}& {32.5} & {27.5} & 23.1 & 18.8 & 14.7 & 11.0 & 7.7 & 5.2 & 3.1 \\
    & SmoothAdv  & 0.790 & 43.7 & 40.3 & 36.9 & 33.8 & 30.5 & 27.0 & 24.0 & 21.4 & 18.4 & 15.9 & 13.4 \\
    & MACER  & 0.744 & {41.4} & 38.5 & 35.2 & 32.3 & 29.3 & 26.4 & 23.4 & 20.2 & 17.4 & 14.5 & 12.1 \\
    & Consistency  & 0.756 & 46.3 & \underline{42.2} & \underline{38.1} & \underline{34.3} & 30.0 & 26.3 & 22.9 & 19.7 & 16.6 & 13.8 & 11.3 \\ 
    & SmoothMix  & 0.773 & 45.1 & 41.5 & 37.5 & 33.8 & 30.2 & 26.7 & 23.4 & 20.2 & 17.2 & 14.7 & 12.1 \\
    \cmidrule(l){2-2} \cmidrule(l){3-3} \cmidrule(l){4-14}
    & \textbf{CAT-RS (Ours)} & \underline{\textbf{0.815}} & 43.2 & {40.2} & \textbf{37.2} & \underline{\textbf{34.3}} & \underline{\textbf{31.0}} & \underline{\textbf{28.1}} & \underline{\textbf{24.9}} & \underline{\textbf{22.0}} & \underline{\textbf{19.3}} & \underline{\textbf{16.8}} & \underline{\textbf{14.2}}\\  
    
    \bottomrule
\end{tabular}
\end{adjustbox}
\caption{Comparison of ACR and approximate certified test accuracy (\%) on CIFAR-10. For each column, we set our result bold-faced if it improves the Gaussian baseline. We set the result underlined if it achieves the highest among the baselines.}
\label{tab:cifar10}
\end{table*}

\begin{table*}[ht]
\centering
\small

\begin{tabular}{lc|cccccccc}
    \toprule
    Methods & ACR & 0.0 & 0.5 & 1.0 & 1.5 & 2.0 & 2.5 & 3.0 & 3.5 \\ 
    \midrule
    Gaussian & 0.875 & {44} & 38 & 33 & 26 & 19 & 15 & 12 & 9 \\
    Consistency & 0.982 & 41 & 37 & 32 & 28 & 24 & 21 & 17 & 14 \\ 
    SmoothAdv & 1.040 & 40 & 37 & 34 & 30 & 27 & 25 & 20 & 15 \\
    SmoothMix & 1.047 & 40 & 37 & 34 & 30 & 26 & 24 & 20 & 17 \\
    \cmidrule(l){1-1} \cmidrule(l){2-2} \cmidrule(l){3-10}
    \textbf{CAT-RS (Ours)} & \underline{\textbf{1.071}} & \underline{\textbf{44}} & \underline{\textbf{38}} & \underline{\textbf{35}} & \underline{\textbf{31}} & \underline{\textbf{27}} & \underline{\textbf{24}} & \underline{\textbf{20}} & \underline{\textbf{17}}\\ 
    \bottomrule
\end{tabular}

\caption{Comparison of ACR and approximate certified accuracy (\%) on ImageNet. For each column, we set our result bold-faced whenever it improves the Gaussian baseline. We set the result underlined if it achieves the highest among the baselines.}
\label{tab:imagenet}
\end{table*}

\section{Experiments}
\label{s:experiments}

\begin{table*}[ht]
\begin{minipage}{.5\textwidth}
\centering
\begin{adjustbox}{height=1.4in}
\begin{tabular}{l|cccccc|c}
    Type & \rot{Gaussian} & \rot{Stability} & \rot{SmoothAdv} & \rot{MACER} & \rot{Consistency} & \rot{SmoothMix} & \rot{\textbf{CAT-RS (Ours)}} \\ 
    \midrule
    Gaussian & 0.412 & 0.348 & 0.506 & 0.473 & {0.505} & \underline{0.513} & \textbf{0.544}\\
    Shot & 0.414 & 0.350 & 0.503 & 0.472 & {0.503} & \underline{0.508} & \textbf{0.542}\\
    Impulse & 0.389 & 0.322 & {0.495} & 0.452 & 0.492 & \underline{0.499} & \textbf{0.530}\\ 
    Defocus & 0.372 & 0.329 & {0.480} & 0.442 & 0.482 & \underline{0.489} & \textbf{0.512}\\
    Glass & 0.343 & 0.291 & {0.473} & 0.415 & 0.472 & \underline{0.483} & \textbf{0.505} \\
    Motion & 0.352 & 0.314 & 0.458 & 0.417 & {0.465} & \underline{0.474} & \textbf{0.492}\\
    Zoom & 0.346 & 0.315 & {0.468} & 0.420 & 0.462 & \underline{0.476} & \textbf{0.501}\\
    Snow & 0.346 & 0.325 & \underline{0.452} & 0.417 & {0.448} & 0.438 & \textbf{0.487}\\
    Frost & 0.298 & 0.298 & \textbf{0.434} & 0.377 & 0.401 & 0.403 & \textbf{0.434} \\
    Fog & 0.197 & 0.153 &\underline{0.279} & 0.266 & {0.277} & 0.262 & \textbf{0.293}\\
    Bright & 0.378 & 0.366 & 0.487 & 0.451 & \underline{0.489} & 0.478 & \textbf{0.524}\\
    Constrast & 0.146 & 0.131 & \textbf{0.228} & 0.195 & 0.213 & 0.202 & \textbf{0.228}\\
    Elastic & 0.331 & 0.290 & 0.441 & 0.405 & {0.445} & \underline{0.447} & \textbf{0.464}\\
    Pixel & 0.404 & 0.350 & 0.500 & 0.465 & {0.500} & \underline{0.509} & \textbf{0.538}\\
    JPEG & 0.413 & 0.354 & \underline{0.504} & 0.470 & 0.502 & \underline{0.504} & \textbf{0.537}\\
    \midrule 
    \textbf{mACR} & 0.343 & 0.302 & \underline{0.447} & 0.409 & 0.444 & 0.446 & \textbf{0.475}\\
    \bottomrule
\end{tabular}
\end{adjustbox}
\caption{Comparison of \emph{average certified radius} (ACR) on CIFAR-10-C. We report the average across five different corruption severities. We set the highest and runner-up values bold-faced and underlined, respectively.}
\label{tab:cifar10c_rotated}
\end{minipage}
\hspace*{\fill}
\begin{minipage}{.45\textwidth}
\centering
\begin{adjustbox}{height=1.4in}
\begin{tabular}{l|cccccc|c}
    Type & \rot{Gaussian} & \rot{Stability} & \rot{SmoothAdv} & \rot{MACER} & \rot{Consistency } & \rot{SmoothMix} & \rot{\textbf{CAT-RS (Ours)}} \\ 
    \midrule
    Clean & 76.6 & 73.0 & 73.4 & \textbf{79.5} & 75.8 & {77.1} & 76.3 \\
    \midrule
    Gaussian & 70.8 & 64.6 & 70.2 & {72.6} & 69.8 & \underline{73.4} & \textbf{76.8}\\
    Shot & 70.0 & 65.6 & 68.4 & \underline{72.8} & 69.6 & 72.6 & \textbf{76.6}\\
    Impulse & 70.2 & 61.6 & 69.0 & \underline{74.0} & 70.4 & 73.6 & \textbf{75.6}\\ 
    Defocus & 64.8 & 65.4 & 68.4 & \underline{71.2} & 69.2 & 70.6 & \textbf{74.2}\\
    Glass & 65.2 & 62.0 & 68.6 & {71.6} & 69.0 & \underline{72.0} & \textbf{72.8} \\
    Motion & 66.2 & 62.4 & 67.2 & \textbf{72.2} & 70.8 & 69.6 & \underline{71.6}\\
    Zoom & 65.2 & 64.2 & 65.6 & 70.6 & {68.4} & \underline{71.4} & \textbf{75.4}\\
    Snow & 67.0 & 64.6 & 64.0 & \underline{70.8} & 67.0 & 69.2 & \textbf{71.4}\\
    Frost & 65.6 & 63.0 & 64.0 & \underline{69.0} & 66.8 & \textbf{70.2} & {67.8} \\
    Fog & 52.4 & 38.8 & 45.4 & \textbf{53.8} & 49.2 & {50.4} & \underline{51.4}\\
    Bright & 71.0 & 70.6 & 67.6 & \underline{73.8} & {73.2} & \underline{73.8} & \textbf{76.4}\\
    Constrast & {39.4} & 30.0 & 34.8 & \textbf{42.8} & 35.6 & 36.4 & \underline{37.8}\\
    Elastic & 64.4 & 63.4 & 64.6 & \underline{71.0} & 66.4 & 69.8 & \textbf{71.4}\\
    Pixel & 66.4 & 67.6 & 68.6 & \underline{74.4} & 69.8 & 69.8 & \textbf{76.2}\\
    JPEG & 67.8 & 66.8 & 68.6 &\underline{70.8} & {68.4} & \underline{70.8} & \textbf{76.2}\\
    \midrule 
    \textbf{mAcc} & 64.4 & 60.7 & 63.7 & \underline{68.8} & 65.6 & 67.7 & \textbf{70.1}\\
    \bottomrule
\end{tabular}
\end{adjustbox}
\caption{Comparison of certified accuracy at $r=0.0$ (\%) on CIFAR-10-C. We report the average across five different corruption severities. We set the highest and runner-up values bold-faced and underlined, respectively.}
\label{tab:cifar10c_clean_rotated}
\end{minipage}
\hspace*{\fill}
\end{table*}

We evaluate the effectiveness of our proposed training scheme based on various well-established image classification benchmarks to measure robustness, including MNIST \cite{dataset/mnist}, Fashion-MNIST \cite{xiao2017fashion}, CIFAR-10/100 \cite{dataset/cifar}, and {ImageNet} \cite{dataset/ilsvrc} (for certified robustness)\footnote{Results on {MNIST, Fashion-MNIST, and CIFAR-100} can be found in Appendix~\ref{ap:addition_dataset}.}, as well as MNIST-C \cite{mu2019mnistc}\footnote{Results on {MNIST-C} can be found in Appendix~\ref{ap:mnistc}.} and CIFAR-10-C \cite{hendrycks2018benchmarking} (for corruption robustness). For a fair comparison, we follow the standard protocol and training setup of the previous works \cite{pmlr-v97-cohen19c,Zhai2020MACER,jeong2020consistency}.\footnote{More details, \eg training setups, datasets, and hyperparameters, can be found in Appendix~\ref{experiment:detail}.} 

Overall, the results show that our method can consistently outperform the previous best efforts to improve the average certified radius by (a) maximizing the robust radii of high-confidence samples while (b) better maintaining the accuracy at low-confidence samples.\footnote{Although our experiments are mainly based on $\ell_2$, we also provide results for $\ell_\infty$ adversary on CIFAR-10 in Appendix~\ref{ap:curves}.} Moreover, the results on CIFAR-10-C, a corrupted version of CIFAR-10, show that our training scheme also helps smoothed classifiers to generalize on out-of-distribution inputs beyond adversarial examples, as shown by a significant improvement in corruption robustness compared to other robust training methods. We also perform an ablation study, showing that, \eg the hyperparameter $\lambda$ in \eqref{eq:overall} between $L^{\tt low}$ and $L^{\tt high}$ can balance the trade-off between robustness and accuracy well.

\paragraph{Baselines.} We compare our method with an extensive list of baseline methods in the literature of training smoothed classifiers:\footnote{We do not compare with empirical defenses such as adversarial training \cite{madry2018towards} as they cannot provide robustness certification: instead, we do compare with SmoothAdv \cite{nips_salman19} that adopts adversarial training for smoothed classifiers.} (a) \emph{Gaussian training} \cite{pmlr-v97-cohen19c} simply trains a classifier with Gaussian augmentation \eqref{eq:gaussian_training}; (b) \emph{Stability training} \cite{li2019stab} adds a cross-entropy term between the logits from clean and noisy images; (c) \emph{SmoothAdv} \cite{nips_salman19} employs adversarial training for smoothed classifiers \eqref{eq:smoothadv}; (d) \emph{MACER} \cite{Zhai2020MACER} adds a regularization that aims to maximize a soft approximation of certified radius; (e) \emph{Consistency} \cite{jeong2020consistency} regularizes the variance of confidences over Gaussian noise; (f) \emph{SmoothMix} \cite{jeong2021smoothmix} proposes a mixup-based \cite{zhang2018mixup} adversarial training for smoothed classifiers. Whenever possible, we use the pre-trained models publicly released by the authors to reproduce the results.

\paragraph{Evaluation metrics.} We follow the standard evaluation protocol for smoothed classifiers \cite{nips_salman19,Zhai2020MACER,jeong2020consistency,jeong2021smoothmix}: specifically, \citet{pmlr-v97-cohen19c} has proposed a practical Monte-Carlo-based certification procedure, namely \textsc{Certify}, that returns the prediction of $\hat{f}$ and a lower bound of certified radius, $\mathrm{CR}(f, \sigma, x)$, over the randomness of $n$ samples with probability at least $1-\alpha$, or abstains the certification. Based on \textsc{Certify}, we consider two major evaluation metrics: (a) the \emph{average certified radius} (ACR) \cite{Zhai2020MACER}: the average of certified radii on the test set $\mathcal{D}_{\tt test}$ while assigning incorrect samples as 0: 
\begin{equation}
\mathrm{ACR} := \frac{1}{|\mathcal{D}_{\tt test}|}\sum_{(x, y)\in \mathcal{D}_{\tt test}}[\mathrm{CR}(f, \sigma, x) \cdot \mathds{1}_{\hat{f}(x)=y}],
\end{equation}
and (b) the \emph{approximate certified test accuracy} at $r$: the fraction of the test set which \textsc{Certify} classifies correctly with the radius larger {than $r$} without abstaining. We use $n=100,000$, $n_0=100$, and $\alpha=0.001$ for \textsc{Certify}, following previous works \cite{pmlr-v97-cohen19c,nips_salman19,jeong2020consistency,jeong2021smoothmix}. 

\subsection{Results on CIFAR-10} 
Table~\ref{tab:cifar10} shows the performance of the baselines and our model on CIFAR-10 for $\sigma \in \{0.25,0.5,1.0\}$. We also plot the approximate certified accuracy over $r$ in Figure~\ref{fig:cifar10} (of Appendix~\ref{ap:curves}). For the baselines, we report best-performing configurations for each $\sigma$ in terms of ACR among reported in previous works, so that the hyperparameters of the same method can vary over $\sigma$ (the details can be found in Appendix~\ref{ap:hyperparameters}). Overall, CAT-RS achieves a significant improvement of ACR compared to the baselines. In case of $\sigma=0.25$ and $\sigma=0.5$, CAT-RS clearly offers a better trade-off between the clean accuracy and robustness compared to other baselines. Especially, CAT-RS achieves higher approximate certified accuracy for all radii compared to SmoothMix in case of $\sigma=0.5$. For $\sigma=1.0$, the ACR of our method significantly surpasses the previous best model, SmoothMix, by $0.773 \rightarrow 0.815$. The improvement of CAT-RS is most evident in $\sigma=1.0$. This means that our proposed CAT-RS can be more effective at challenging tasks, where it is more likely that a given classifier gets a more diverse confidence distribution for the training samples, so that our proposed confidence-aware training can better play its role.

\begin{figure*}[t]
\centering
\begin{minipage}{.64\textwidth}
    \centering
    \subfigure[Effect of $\lambda$]
    {
        \includegraphics[width=0.44\linewidth]{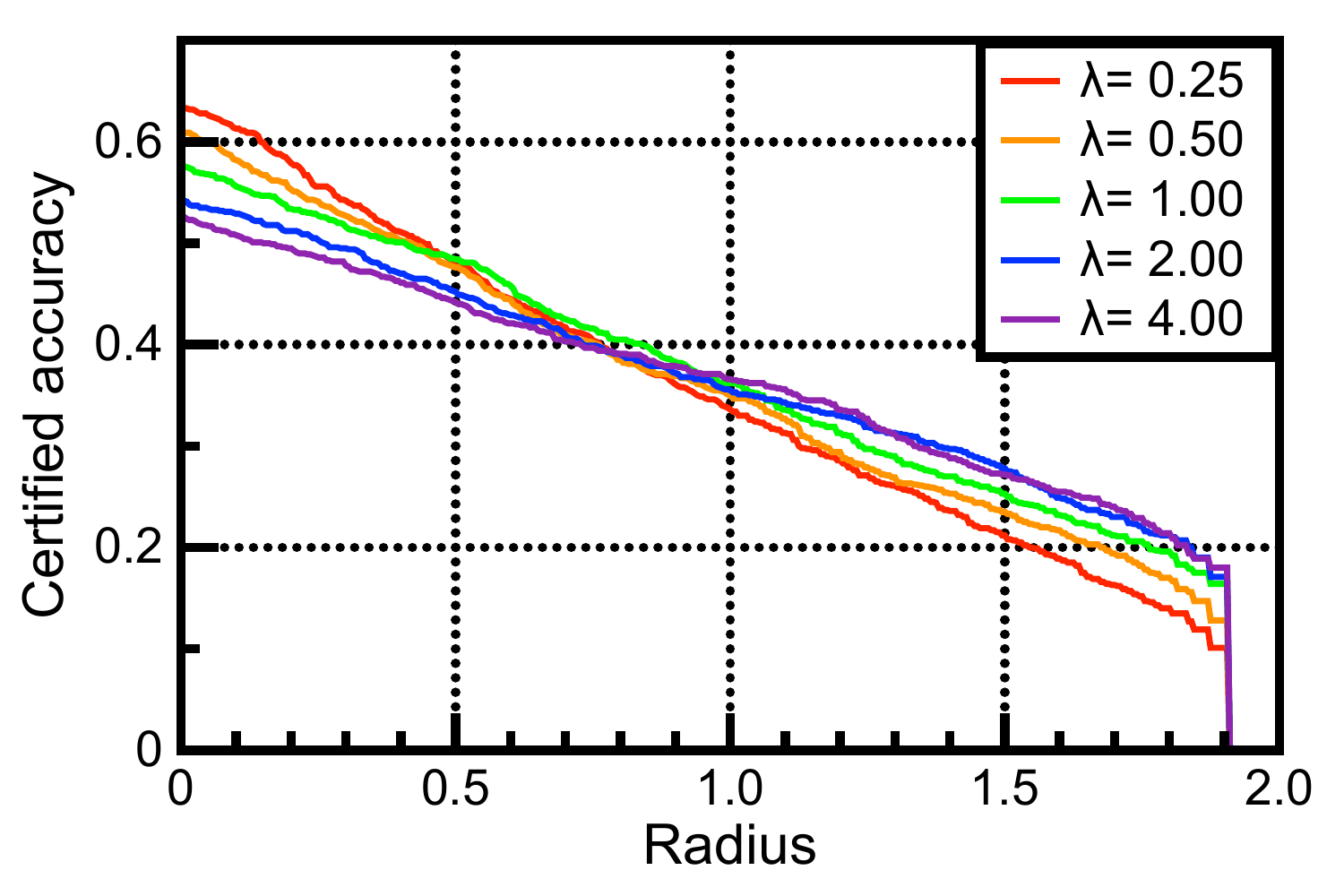}
    	\label{fig:abla_lbd}
    }
    \subfigure[Effect of $M$]
    {
        \includegraphics[width=0.44\linewidth]{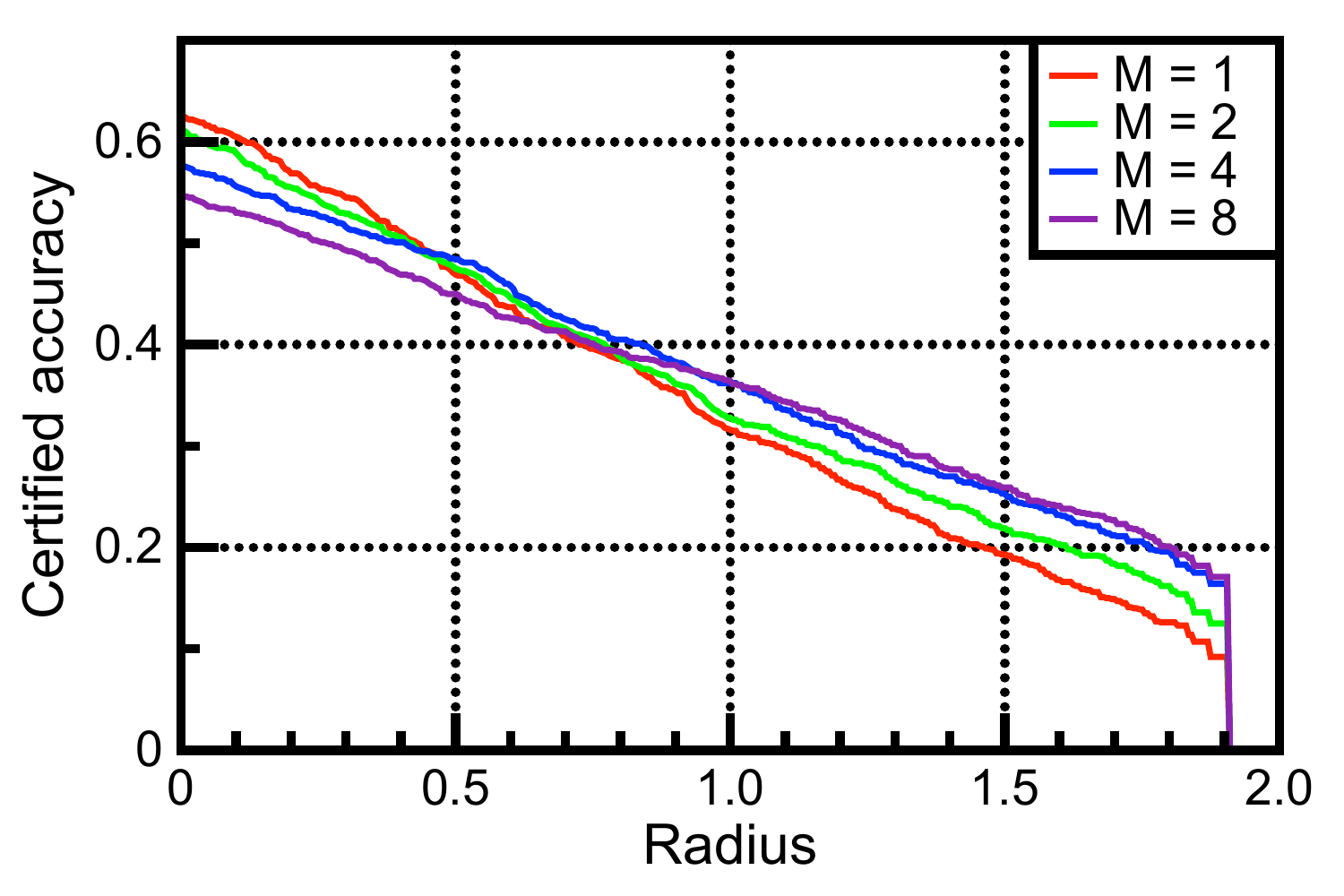}
    	\label{fig:abla_M}
    }
    \caption{Comparison of certified accuracy of CAT-RS ablations on CIFAR-10. We use ResNet-20 for ablation study and plot the results at $\sigma=0.5$. Detailed results on ablation experiments can be found in Appendix~\ref{ap:abla}.}
    \label{fig:abla}
\end{minipage}
\hfill
\begin{minipage}{.33\textwidth}
  \centering
	\includegraphics[width=0.9\linewidth]{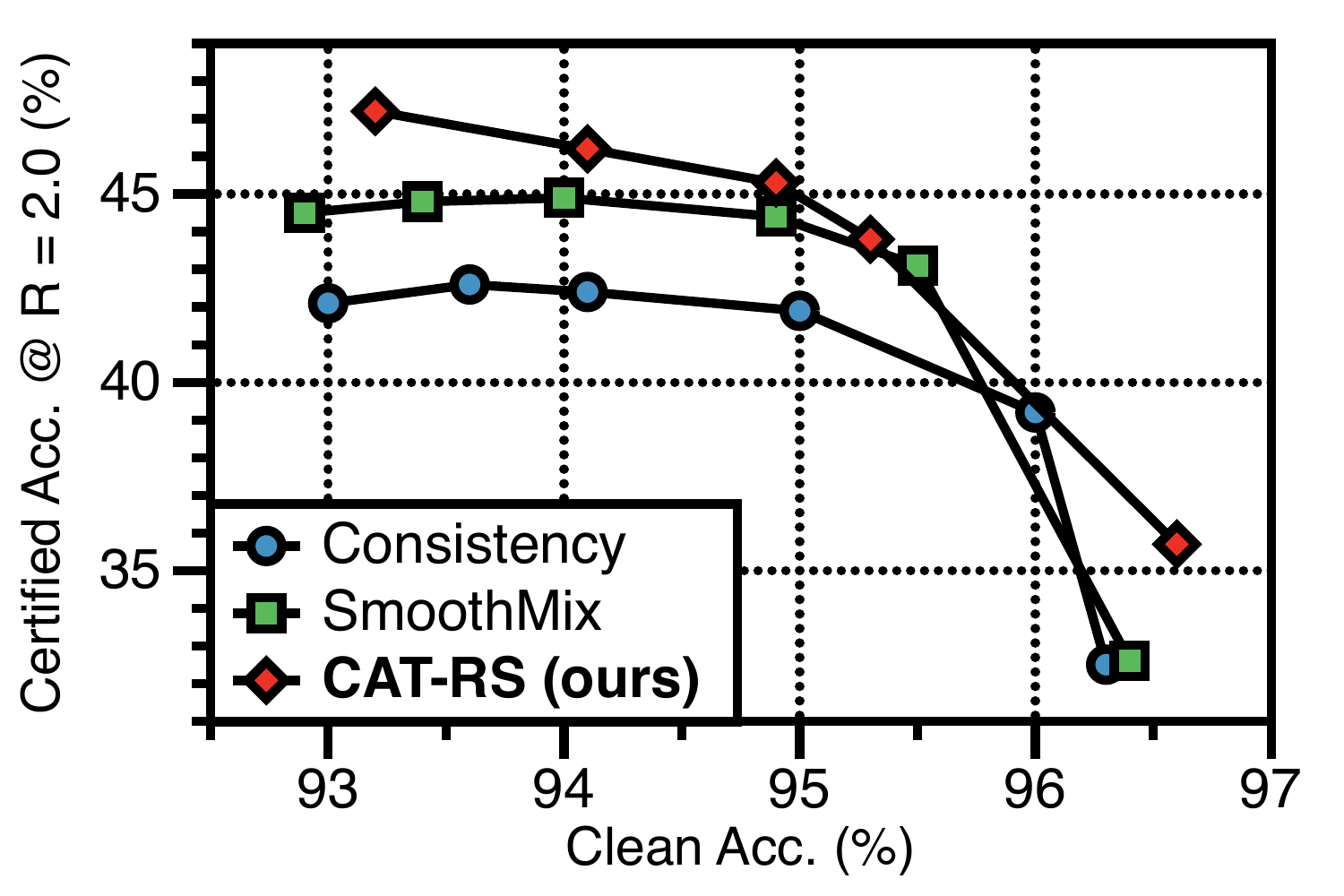}
	\vspace{0.15in}
	\caption{Trade-off between clean \emph{vs.}~certified acc. on MNIST ($\sigma=1.0$) for varying control hyperparameter.}
	\label{fig:acr_clean}
\end{minipage}
\end{figure*}

\subsection{Results on ImageNet}

In this section, we compare the certified robustness of our method on ImageNet \cite{dataset/ilsvrc} dataset for $\sigma=1.0$. We evaluate the performance on the uniformly-subsampled 500 samples in the ImageNet validation dataset following \cite{pmlr-v97-cohen19c,jeong2020consistency,nips_salman19,jeong2021smoothmix}. The results shown in Table~\ref{tab:imagenet} confirm that our method achieves the best results in terms of ACR and certified test accuracy compared to the considered baselines, verifying the effectiveness of CAT-RS even in the large-scale dataset.

\subsection{Results on CIFAR-10-C}

We also examine the performance of CAT-RS on CIFAR-10-C \cite{hendrycks2018benchmarking}, a collection of 75 replicas of the CIFAR-10 test dataset, which consists of 15 different types of common corruptions (\eg fog, snow, etc.), each of which contains 5 levels of corruption severities. Similarly to \cite{sun2021certified}, for a given smoothed classifier trained on CIFAR-10, we report ACR and the certified accuracy at $r=0.0$ for each corruption type of CIFAR-10-C after averaging over five severity levels, as well as their means over the types, \ie as the \emph{mean-ACR} (mACR) and \emph{mean-accuracy} (mAcc), respectively. We uniformly subsample each corrupted dataset with size 100, \ie to have 7,500 samples in total, and use $\sigma=0.25$ throughout this experiment.  

Table~\ref{tab:cifar10c_rotated} and \ref{tab:cifar10c_clean_rotated} summarizes the results. Overall, CAT-RS achieves the best ACRs on all the corruption types, thus also in mACR, as well as it significantly improves mAcc compared to other methods, \ie for 11 out of 15 corruption types. In other words, CAT-RS can improve smoothed classifiers to generalize better on unseen corruptions, at the same time maintaining the robustness for such inputs. It is remarkable that the observed gains are not from any prior knowledge about multiple corruption \cite{hendrycks2020augmix,Hendrycks_2021_ICCV} (except for Gaussian noise), but from a better training method. Given the limited gains from other baseline methods on CIFAR-10-C, we attribute that the \emph{sample-dependent calibration} of training objective, a unique aspect of CAT-RS compared to prior arts, is important to explain the effectiveness of CAT-RS on out-of-distribution generalization: \eg although SmoothAdv also adopts adversarial search in training similarly to CAT-RS, it could not improve mAcc on CIFAR-10-C from Gaussian.

\subsection{Ablation Study}

In this section, we conduct an ablation study to further analyze individual effectiveness of the design components in our method. Unless otherwise specified, we use ResNet-20 \cite{he2016deep} and test it on a uniformly subsampled CIFAR-10 test set of size 1,000. We provide more ablations on the loss design and the detailed results in Appendix~\ref{ablation:detail}.

\paragraph{Effect of $\lambda$.} 

In CAT-RS, $\lambda$ introduced in \eqref{eq:overall} controls the relative contribution of $L^{\mathtt{high}}$ over $L^{\mathtt{low}}$. Here, Figure~\ref{fig:abla_lbd} shows the impact of $\lambda$ to the model on varying $\lambda \in \{0.25,0.5,1.0,2.0,4.0\}$, assuming $\sigma=0.5$. The results show that $\lambda$ successfully balances the trade-off between robustness and clean accuracy \cite{pmlr-v97-zhang19p}. In addition, Figure~\ref{fig:acr_clean} further verifies that CAT-RS offers more effective trade-off compared to other baseline training methods, as further discussed later in this section. 

\paragraph{Effect of $M$.} 

We investigate the effect of the number of noise $M$. Figure~\ref{fig:abla_M} illustrates the approximate test certified accuracy with varying $M \in \{1,2,4,8\}$. The robustness of the smoothed classifier increases as $M$ increases, sacrificing its clean accuracy. For large $M$, the classifier can incorporate the information of many Gaussian noises and take advantage of increasing ${p}_f$ \eqref{eq:p_f}. Therefore, the smoothed classifier can provide a more robust prediction.

\paragraph{Accuracy-robustness trade-off.}

To further validate that our method can exhibit a better trade-off between accuracy and robustness compared to other methods, we additionally compare the performance trends between clean accuracy and certified accuracy at $r=2.0$ as we vary a hyperparameter to control the trade-off, \eg $\lambda$ \eqref{eq:overall} in case of our method. We use $\sigma=1.0$ {on MNIST dataset} for this experiment. We choose Consistency and SmoothMix for this comparison, considering that they also offer a single hyperparameter (namely $\lambda$ and $\eta$, respectively) for the balance between accuracy and robustness similar to our method, while both generally achieve good performances among the baselines considered. The results plotted in Figure~\ref{fig:acr_clean} show that CAT-RS indeed exhibits a higher trade-off frontier compared to both methods, which confirms the effectiveness of our method. More detailed results can be found in Appendix~\ref{ap:trade-off}.

\section{Conclusion}

This paper explores a close relationship between confidence and robustness, a natural property of smoothed classifiers yet neural networks cannot currently offer. We have successfully leveraged this to relax the hard-to-compute metric of adversarial robustness into an easier concept of prediction confidence. Consequently, we propose a practical training method that enables a sample-level control of adversarial robustness, which has been difficult in a conventional belief. We believe our work could be a useful step for the future research on exploring the interesting connection between adversarial robustness and \emph{confidence calibration} \cite{pmlr-v70-guo17a}, and even towards the \emph{out-of-distribution generalization}, through the randomized smoothing framework.

\section*{Acknowledgments}
This work was conducted by Center for Applied Research in Artificial Intelligence (CARAI) grant funded by Defense Acquisition Program Administration (DAPA) and Agency for Defense Development (ADD) (UD190031RD). 

\bibliography{aaai23}

\clearpage
\onecolumn
\appendix


\begin{center}
{\bf {\LARGE Supplementary Material}} \\
\vspace{0.05in}
{\bf {\Large Confidence-aware Training of Smoothed Classifiers for Certified Robustness}}
\end{center}

\section{Training procedure of CAT-RS}
\label{ap:alg}

\begin{algorithm}[ht]
\caption{Confidence-aware Training for Randomized Smoothing (CAT-RS)}\label{alg:training}
\begin{algorithmic}[1]
\REQUIRE training sample $(x, y)$. smoothing factor $\sigma$. number of noise samples $M$. consistency targets $\hat{y}\in\Delta^{K-1}$, regularization strength $\lambda > 0$. attack norm $\varepsilon > 0$.
\vspace{0.05in}
\hrule
\vspace{0.05in}
\STATE Sample $\delta_1, \cdots, \delta_M \sim \mathcal{N}(0, \sigma^2 I)$
\STATE $\hat{p}_f \leftarrow \frac{1}{M}\sum_i \mathds{1}[f(x+\delta_i)=y]$
\STATE Sample $K \sim \mathrm{Bin}(M, \hat{p}_f), K^+ \leftarrow \max(1, K)$
\FOR{$i=1$ {\bfseries to} $M$}
    \STATE $L_i \leftarrow \mathbb{CE}(F(x+\delta_i), y)$
    \STATE $\delta^*_i \leftarrow \argmax_{\|\delta^*_i - \delta_i \|\le \varepsilon} \mathrm{KL}(F(x+\delta^*_i), \hat{y})$
    
\ENDFOR
\STATE $L^{\pi}_{1:M} \leftarrow \mathtt{argsort}(L_{1:M})$
\STATE $L^{\tt low}, L^{\tt high} \leftarrow \frac{1}{M}(\sum_{i=1}^{K^+} L^{\pi}_{i}),\  \max_i \mathrm{KL}(F(x+\delta^*_i), \hat{y})$
\STATE $L^{\mathtt{CAT}\text{-}\mathtt{RS}} \leftarrow 
L^{\tt low} + \lambda \cdot \mathds{1}[K^+ = M] \cdot  L^{\tt high}$
\end{algorithmic}
\end{algorithm}

\section{Experimental details}
\label{experiment:detail}

We follow the training setup considered in most of the previous works to compare the performance of the smoothed classifiers \cite{pmlr-v97-cohen19c,Zhai2020MACER,jeong2020consistency,jeong2021smoothmix}: specifically, we mainly consider LeNet \cite{dataset/mnist}, ResNet-110 \cite{he2016deep}, and ResNet-50 for MNIST{/Fashion-MNIST}, CIFAR-10/100, and ImageNet, respectively, and consider different scenarios of $\sigma \in \{0.25, 0.5, 1.0$\} for randomized smoothing. We apply the same $\sigma$ for both training and evaluation. When training, we use stochastic gradient descent (SGD) optimizer with a momentum of 0.9, and weight decay of $10^{-4}$. The learning rate is initialized to 0.01 for MNIST{/Fashion-MNIST} and 0.1 for CIFAR-10/100, and decreased by a factor of 0.1 in every 50 epochs within 150 training epochs. {For ImageNet, we train ResNet-50 \cite{he2016deep} for 90 epochs, with the initial learning rate of 0.1 decreased by a factor of 0.1 in every 30 epochs, additionally by a factor of 0.1 for the last 5 epochs. We use $\varepsilon=1.0$ for 80 epochs of training and increase it to $\varepsilon=2.0$ for the last 10 epochs. Also, to further alleviate the cold-start problem in \eqref{eq:bottom_k} under many-class ImageNet, we assume $K\sim \mathrm{Bin}(M, \hat{y}_c)$ instead of $K\sim \mathrm{Bin}(M, \hat{p}_f(x, y))$ so that the training can avoid binomial sampling from $\hat{p}_f(x,y) \approx 1/C$ for the early stage of training.}

\subsection{Datasets}
\label{ap:datasets}

\paragraph{MNIST} \citep{dataset/mnist} consists of 70,000 gray-scale hand-written digit images of size 28$\times$28, 60,000 for training and 10,000 for testing, where each is labeled to one value between 0 and 9. We do not perform any pre-processing except for normalizing the range of each pixel from 0-255 to 0-1. The dataset can be downloaded at \url{http://yann.lecun.com/exdb/mnist/}. 

\paragraph{Fashion-MNIST} \citep{xiao2017fashion} {consists of 70,000 gray-scale 10-category fashion product images of size $28 \times 28$, 60,000 for training and 10,000 for testing. Each category is assigned to one value between 0 and 9, where each image is labeled to the value assigned to its category. We do not perform any pre-processing except for normalizing the range of each pixel from 0-255 to 0-1. The dataset can be downloaded at \url{https://github.com/zalandoresearch/fashion-mnist}.}

\paragraph{CIFAR-10/100} \citep{dataset/cifar} consists of 60,000 RGB images of size 32$\times$32, 50,000 for training and 10,000 for testing, where each is labeled to one of 10 and 100 classes, respectively. We use the standard data-augmentation scheme of random horizontal flip and random translation up to 4 pixels, following the practice of other baselines \cite{pmlr-v97-cohen19c,nips_salman19,Zhai2020MACER,jeong2020consistency,jeong2021smoothmix}. We also normalize the images in pixel-wise by the mean and the standard deviation calculated from the training set. The full dataset can be downloaded at \url{https://www.cs.toronto.edu/~kriz/cifar.html}.

\paragraph{ImageNet} \citep{dataset/ilsvrc} consists of 1,281,167 images for training, and 50,000 images for validation. Each of the images are labeled to one of 1,000 classes. We perform 224$\times$224 randomly resized cropping and horizontal flipping for the training images. For test images, we resize the images into 256$\times$256 resolution, followed by 224$\times$224 center cropping. The full dataset can be downloaded at \url{https://image-net.org/download}.

\subsection{Hyperparameters}
\label{ap:hyperparameters}

\paragraph{Stability training} \cite{li2019stab} {introduces} a single hyperparameter $\gamma$ to control the relative strength of the regularization {for the logits under Gaussian augmentation}. We fix $\gamma=2$ for MNIST{/Fashion-MNIST}. For CIFAR-10/100, $\gamma=2$ is used for $\sigma=0.25, 0.5$, and $\gamma=1$ is used for $\sigma=1.0$.

\paragraph{SmoothAdv} \cite{nips_salman19} uses three major hyperparameters to perform the projected gradient descent: namely, the attack radius in terms of  $\ell_2$-norm $\varepsilon$, the number of PGD steps $T$, and the number of noises $m$. In our experiments, we fix $T=10$. For MNIST{/Fashion-MNIST}, we fix $\varepsilon=1.0$ and $m=4$ as well. In case of CIFAR-10/100, on the other hand, we report the results chosen among the list of ``best'' configurations for each noise level which are previously searched by \citet{nips_salman19}: specifically, we report the results of $\varepsilon=1.0$ and $m=4$ for $\sigma=0.25$, and $\varepsilon=1.0$ and $m=8$ for $\sigma=0.5$, and $\varepsilon=2.0$ and $m=2$ for $\sigma = 1.0$. When SmoothAdv is used, we adopt the \emph{warm-up} strategy, \ie we initially set $\varepsilon=0.0$ and linearly increase to the target value of $\varepsilon$ for 10-epochs.

\paragraph{MACER} \cite{Zhai2020MACER} introduces four hyperparameters: the number of noises $k$, the coefficient for the regularization term $\lambda$, the clamping parameter for maximizing the certified radius $\gamma$, and the temperature scaling parameter $\beta$. For MNIST, we use $k=16, \gamma=8.0, \beta=16.0, \text{and } \lambda=16.0$ when $\sigma=0.25, 0.5$, following the configurations in \citet{Zhai2020MACER}. For $\sigma=1.0$, we had to reduce $\lambda=6.0$ for a stable training. {For Fashion-MNIST, we maintain all hyperparameters from MNIST experiments except $\lambda$. For a stable training, we had to set $\lambda=8.0$ and $\lambda=2.0$ for $\sigma=0.5$ and $\sigma=1.0$, respectively.} For CIFAR-10/100, we follow the original configurations used by \citet{Zhai2020MACER}. We set $k=16, \gamma=8.0 $, and $\beta=16.0$. $\lambda$ is set to be 12.0 and 4.0 for $\sigma=0.25$ and $0.5$, respectively. For $\sigma=1.0$, the training starts with $\lambda=0$ until the first learning rate decay and we set $\lambda=12.0$ thereafter.

\paragraph{Consistency} \cite{jeong2020consistency} uses two hyperparameters: namely, the coefficient for the consistency term $\eta$ and the entropy term $\gamma$. We report the best results in terms of ACR among those reported by \citet{jeong2020consistency} varying $\eta$. Following the original practice, we fix $\gamma=0.5$ throughout our experiments. For MNIST{/Fashion-MNIST}, we use $\lambda=10$ for $\sigma=0.25$ and $\lambda=5$ for other noises. For CIFAR-10/100, we use $\lambda=20$ for $\sigma=0.25$ and $\lambda=10$ for other noises.

\paragraph{SmoothMix} \cite{jeong2021smoothmix} introduces four hyperparameters: namely, the mixup {coefficient between the original and adversarial sample} $\eta$, the step size for adversarial attack $\alpha$, the number of steps for adversarial attack $T$, and the number of noises $T$. For MNIST{/Fashion-MNIST}, we fix $\eta=5.0, \alpha=1.0$, and $m=4$. We use $T=2,4,8$ for the models with $\sigma=0.25,0.5,1.0$, respectively. For CIFAR-10/100, we again report the best result among those reported from \citet{jeong2021smoothmix}: \ie we fix $\eta=5.0, m=2,$ and $T=4$, and use $\alpha=0.5, 1.0, 2.0$ for $\sigma=0.25, 0.5, 1.0$, respectively. The ``one-step adversary'' is used for $\sigma=0.5, 1.0$ to follow the best configurations reported.

\paragraph{CAT-RS (Ours)} introduces one main hyperparameter: namely, the coefficient $\lambda$ for the worst-case loss. Although the number of noises $M$, the number of attack steps $T$, and the attack radius $\varepsilon$ are also can be tuned for a better performance, we fix $M=4$, $T=4$, and $\varepsilon=1.0$ unless otherwise noted. For MNIST{/Fashion-MNIST}, we use the fixed configuration of $\lambda=1.0$. For CIFAR-10/100, we use $\lambda=0.5, 1.0, 2.0$ for $\sigma=0.25, 0.5, 1.0$, respectively. For ImageNet, we use $\lambda=2.0$. Also, we set $M=2$ and $T=1$ to reduce the overall training cost. 

For each training sample $x$, we compute its soft-label $\hat{y}$ for \eqref{eq:worst} by the \emph{smoothed prediction} of another classifier $\bar{f}$ pre-trained via Gaussian training \eqref{eq:gaussian_training} with a fixed $\sigma_0=0.25$: specifically, we obtain a soft-label $\hat{y}\in \mathbb{R}^K$ by computing:   
\begin{equation}
\label{eq:smoothed_prediction}
    \hat{y}_c := \frac{1}{N} \sum_{i=1}^N \mathds{1}[\bar{f}(x+\delta_i)=c],
\end{equation}
where $\delta_i \sim \mathcal{N}(0, \sigma_0^2 I)$. In our experiments, we use $N=10,000$ Gaussian noises for MNIST{/Fashion-MNIST} and CIFAR-10/100, and $N=500$ for ImageNet. 

\clearpage
\section{Results on additional datasets}
\label{ap:addition_dataset}

\subsection{Results on MNIST}

\begin{figure*}[ht]
	\centering
	\begin{adjustbox}{width=1.0\linewidth}
	\subfigure[$\sigma=0.25$]
	{
	    \includegraphics[width=0.3\linewidth]{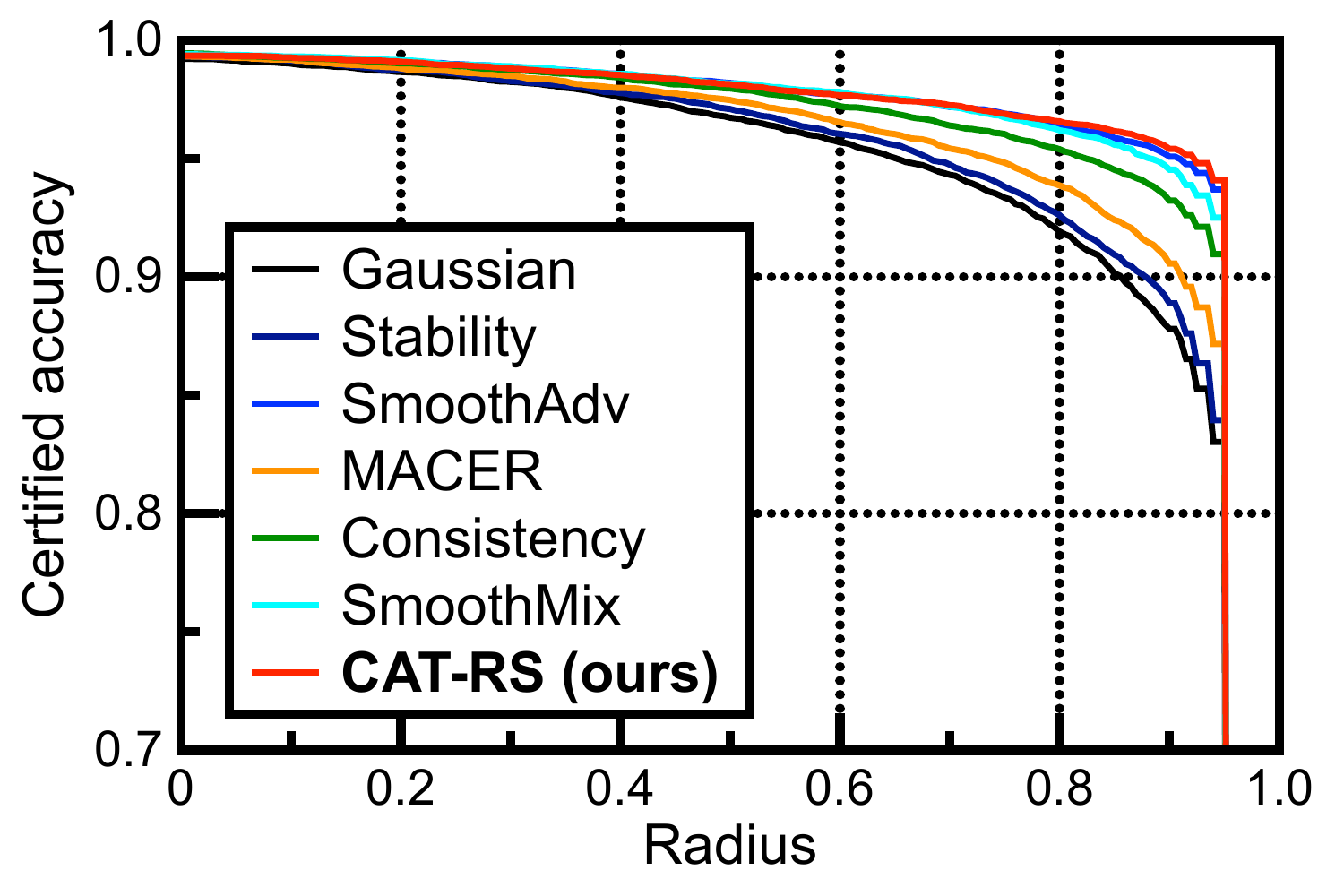}
		\label{fig:mnist_25}
	}
	\subfigure[$\sigma=0.50$]
	{
	    \includegraphics[width=0.3\linewidth]{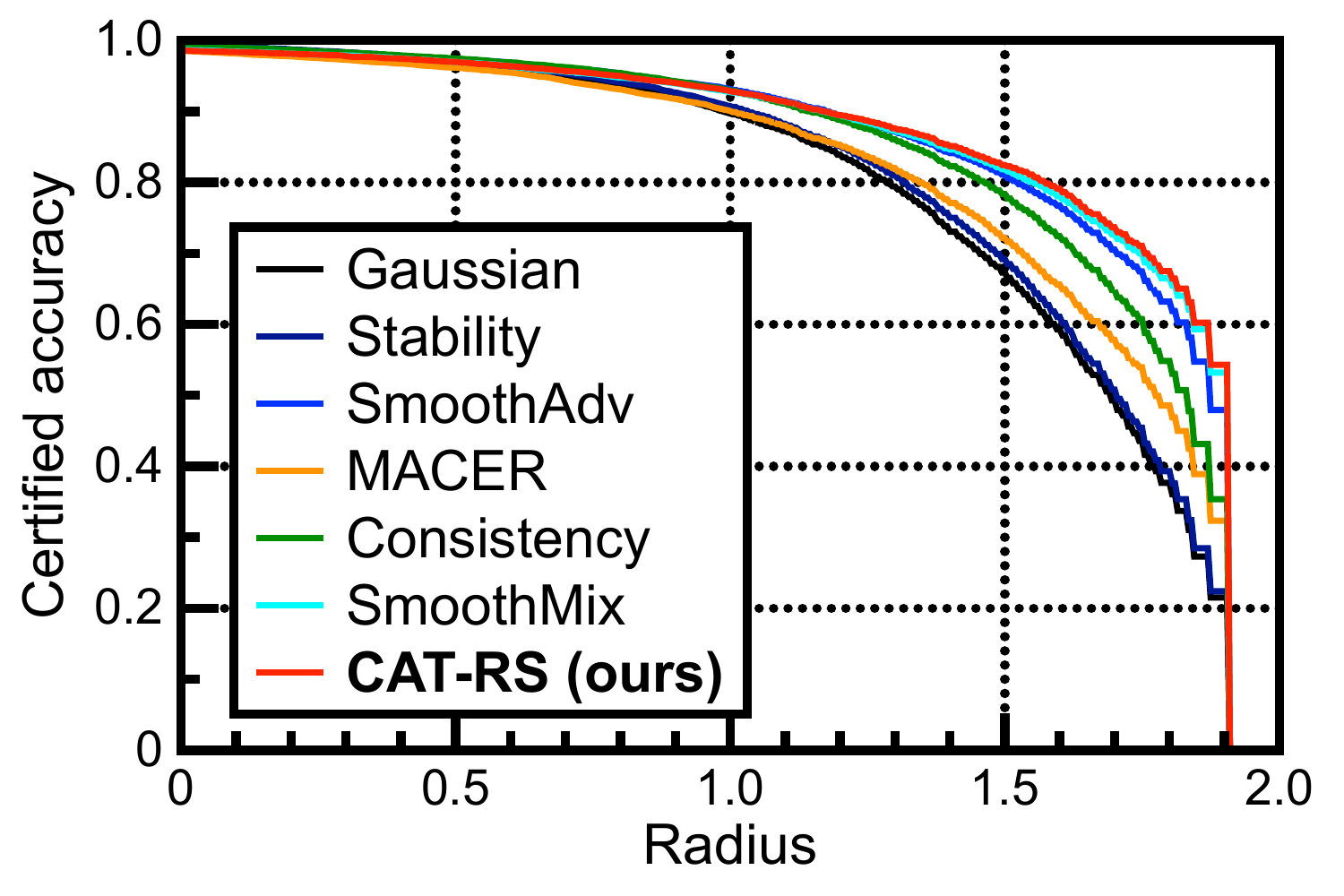}
		\label{fig:mnist_50}
	}
	\subfigure[$\sigma=1.00$]
	{
	    \includegraphics[width=0.3\linewidth]{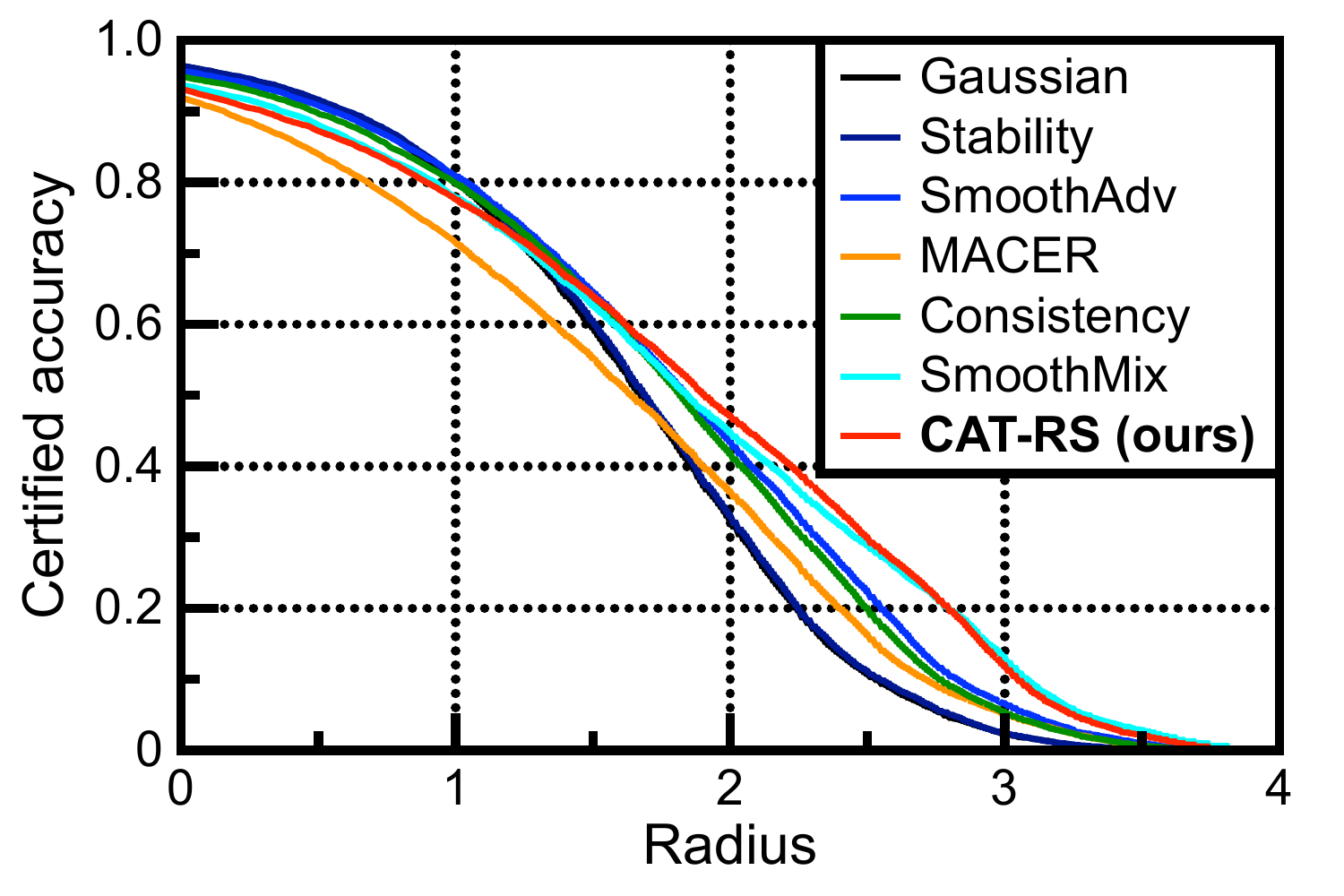}
		\label{fig:mnist_100}
	}
	\end{adjustbox}
	\caption{Comparison of approximate certified accuracy for various training methods on MNIST. The sharp drop of certified accuracy in each plot is due to an upper bound in radius that \textsc{Certify} can output for a given $\sigma$, $N=100,000$, and $\alpha=0.001$.}
	\label{fig:mnist}
\end{figure*}

We compare the certified robustness of the smoothed classifiers trained on MNIST from our method to those from other baselines in Table~\ref{tab:mnist}, considering three different smoothing factors $\sigma\in\{0.25, 0.5, 1.0\}$. We also present in Figure~\ref{fig:mnist} the plots of the approximate certified accuracy across varying $r$. Overall, the results show that CAT-RS clearly surpasses all the other baselines in terms of ACR: \ie our method could better balance between the clean accuracy and robustness. For $\sigma=0.25$, we notice that some baselines, \ie SmoothAdv and SmoothMix, already achieve a reasonably saturated level of ACR: even in this trivial task, our method could further push the boundary of robust accuracies. In more challenging cases of $\sigma=0.5$ and $\sigma=1.0$, on the other hand, the improvements from CAT-RS in ACR become more evident as $\sigma$ increases: \eg at $\sigma=1.0$, compared to SmoothMix (the best-performing baseline), CAT-RS could improve the certified accuracy at $r=2.50$ by $28.9\%\rightarrow 30.0\%$, resulting in ACR increment by 1.820 $\rightarrow$ 1.831. As in CIFAR-10, the improvement of CAT-RS is most evident in $\sigma=1.0$, demonstrating the effectiveness of confidence-aware training.

\begin{table*}[ht]
\centering
\begin{tabular}{clc|ccccccccccc}
    \toprule
    $\sigma$ &  Methods & ACR & 0.00 & 0.25 & 0.50 & 0.75 & 1.00 & 1.25 & 1.50 & 1.75 & 2.00 & 2.25 & 2.50 \\ 
    \midrule
    \multirow{7.5}{*}{0.25}& Gaussian & 0.910 & {99.2} &{98.5} &{96.7} &{{93.3}}& 0.0& 0.0 &0.0 &0.0 &0.0 &0.0 &0.0 \\
    & Stability  & 0.914 & 99.3 &98.6& 97.1& 93.8& 0.0& 0.0& 0.0& 0.0& 0.0& 0.0&0.0 \\ 
    & SmoothAdv  & 0.932 & 99.4 &\underline{99.0}& \underline{98.2}& 96.8& 0.0& 0.0& 0.0 &0.0 &0.0 &0.0 &0.0  \\
    & MACER & 0.921 & 99.3 &98.7& 97.5& 94.8 &0.0 &0.0 &0.0 &0.0& 0.0& 0.0& 0.0 \\  
    & Consistency & 0.928 &\underline{99.5} &98.9 &98.0 &96.0 & 0.0 & 0.0  &0.0 &0.0& 0.0& 0.0& 0.0\\
    & SmoothMix  & 0.932 &  99.4 &\underline{{99.0}}& \underline{98.2}& {{96.7}}& 0.0 & 0.0  &0.0 &0.0& 0.0& 0.0& 0.0\\

    \cmidrule(l){2-2} \cmidrule(l){3-3} \cmidrule(l){4-14}
    & \textbf{CAT-RS (Ours)} & \underline{\textbf{0.933}} & \textbf{99.4} & \underline{\textbf{99.0}} & \underline{\textbf{98.2}} & \underline{\textbf{96.9}} & 0.0 & 0.0 & 0.0 & 0.0 & 0.0 & 0.0 & 0.0\\  
    \midrule
    \multirow{7.5}{*}{0.50}& Gaussian & 1.557 & \underline{99.2} &98.3 &96.8& 94.3& {89.7} &{81.9}& {67.3}& {43.6}& 0.0 & 0.0 & 0.0 \\
    & Stability & 1.573 & \underline{99.2} &98.5 &97.1& 94.8& 90.7& 83.2& 69.2& 45.4&0.0 & 0.0 & 0.0 \\
    & SmoothAdv & 1.687 & 99.0 &98.3 &97.3 &95.8 &\underline{93.2} &88.5 &81.1 &67.5&0.0 & 0.0 & 0.0 \\
    & MACER & 1.583 & {98.5} &{97.5}& {96.2}& {93.7}& 90.0& 83.7& 72.2& 54.0 &0.0 & 0.0 & 0.0\\
    & Consistency & 1.655 & \underline{99.2} &\underline{98.6}& \underline{97.6}& \underline{95.9}& 93.0& 87.8& 78.5& 60.5&0.0 & 0.0 & 0.0\\ 
    & SmoothMix  & 1.694 & 98.7 &98.0 &97.0& 95.3& 92.7& 88.5& 81.8& 70.0&0.0 & 0.0 & 0.0 \\

    \cmidrule(l){2-2} \cmidrule(l){3-3} \cmidrule(l){4-14}
    & \textbf{CAT-RS (Ours)} & \underline{\textbf{1.700}} & 98.6 & 98.0 & \textbf{97.0} & \textbf{95.4} & \textbf{92.8} & \underline{\textbf{88.7}} & \underline{\textbf{82.5}} & \underline{\textbf{71.1}} & 0.0 & 0.0 & 0.0\\  
    \midrule
    \multirow{7.5}{*}{1.00}& Gaussian & 1.619 & 96.3 &94.4& 91.4& 86.8& 79.8& 70.9& 59.4& {46.2}& {32.5}& {19.7}& {10.9} \\
    & Stability & 1.636 & \underline{96.5}& \underline{94.6}& \underline{91.6}& \underline{87.2}& 80.7& 71.7& 60.5& 47.0& 33.4& 20.6& 11.2 \\
    & SmoothAdv & 1.779 & 95.8 &93.9 &90.6& 86.5& \underline{80.8}& \underline{73.7}& \underline{64.6}& 53.9& 43.3& 32.8& 22.2\\
    & MACER & 1.598 & {91.6} &{88.1}& {83.5}& {77.7}& {71.1}& {63.7}& {55.7}& 46.8& 38.4& 29.2& 20.0 \\
    & Consistency & 1.738 & 95.0 &93.0 &89.7& 85.4& 79.7& 72.7& 63.6& 53.0& 41.7& 30.8 &20.3\\ 
    & SmoothMix & 1.820 & 93.7 &91.6 &88.1 &83.5 &77.9 &70.9& 62.7& 53.8& 44.8& 36.6& 28.9 \\
    
    \cmidrule(l){2-2} \cmidrule(l){3-3} \cmidrule(l){4-14}
    & \textbf{CAT-RS (Ours)} & \underline{\textbf{1.831}} & 93.2 & 90.5 & 87.2 & 83.1 & 77.6 & \textbf{71.7} & \textbf{64.0} & \underline{\textbf{55.8}} & \underline{\textbf{47.2}} & \underline{\textbf{39.2}} & \underline{\textbf{30.0}}\\
    \bottomrule
\end{tabular}

\caption{Comparison of ACR and approximate certified test accuracy (\%) on MNIST. For each column, we set our result bold-faced if it improves the Gaussian baseline. We set the result underlined if it achieves the highest among the baselines.}
\label{tab:mnist}
\end{table*}

\subsection{Result on Fashion-MNIST}
\label{ap:FMNIST_datasets}

In this section, we compare the performance on Fashion-MNIST dataset \cite{xiao2017fashion}. Table~\ref{tab:FMNIST} shows ACR and certified accuracy varying the severity of noise level $\sigma \in \{0.25, 0.50, 1.00\}$. Overall, CAT-RS offers a better trade-off between accuracy and robustness, improving ACR compared to the baselines. We highlight that our method is more effective in a challenging setting, \eg $\sigma=1.0$, where leveraging confidence information is critical. For instance, CAT-RS improves the certified accuracy at $r=2.50$ by $28.3\%\rightarrow 31.7\%$, resulting in the increment of ACR by $1.534 \rightarrow 1.607$. It confirms that confidence-aware training can effectively boost the robustness when smoothed via randomized smoothing.

\begin{table*}[ht]
\centering
\begin{tabular}{clc|ccccccccccc}
    \toprule
    $\sigma$ &  Methods & ACR & 0.00 & 0.25 & 0.50 & 0.75 & 1.00 & 1.25 & 1.50 & 1.75 & 2.00 & 2.25 & 2.50\\ 
    \midrule
    \multirow{6.5}{*}{0.25}& Gaussian & 0.670 & \underline{89.5} & 82.0 & 70.8 & 57.7 & 0.0 & 0.0 & 0.0 & 0.0 & 0.0 & 0.0 & 0.0 \\
    & Stability & 0.689 & 89.2 & 83.2 & 73.2 & 60.6 & 0.0 & 0.0 & 0.0 & 0.0 & 0.0 & 0.0 & 0.0\\ 
    & SmoothAdv & 0.756 & 86.2 & 83.3 & \underline{79.8} & 75.1 & 0.0 & 0.0 & 0.0 & 0.0 & 0.0 & 0.0 & 0.0 \\
    & MACER & 0.727 & 88.1 & 84.2 & 77.8 & 68.1 & 0.0 & 0.0 & 0.0 & 0.0 & 0.0 & 0.0 & 0.0 \\  
    & Consistency & 0.744 & 88.5 & \underline{84.7} & 78.8 & 71.2 & 0.0 & 0.0 & 0.0 & 0.0 & 0.0 & 0.0 & 0.0\\
    & SmoothMix & 0.745 & 88.8 & 84.6 & 78.9 & 71.3 & 0.0 & 0.0 & 0.0 & 0.0 & 0.0 & 0.0 & 0.0\\
    \cmidrule(l){2-2} \cmidrule(l){3-3} \cmidrule(l){4-14}
    & \textbf{CAT-RS (Ours)} & \underline{\textbf{0.757}} & 86.3 & \textbf{83.5} & \textbf{79.6} & \underline{\textbf{75.2}} & 0.0 & 0.0 & 0.0 & 0.0 & 0.0 & 0.0 & 0.0\\
    
    \midrule
    \multirow{6.5}{*}{0.50}& Gaussian & 1.056 & \underline{86.2} & 80.7 & 73.2 & 64.8 & 55.5 & 45.6 & 35.0 & 24.1 & 0.0 & 0.0 & 0.0 \\
    & Stability & 1.118 & 85.9 & \underline{81.6} & 75.8 & 68.8 & 60.2 & 50.5 & 39.4 & 27.6 & 0.0 & 0.0 & 0.0 \\
    & SmoothAdv & 1.255 & 83.3 & 80.2 & \underline{76.5} & 71.9 & 66.7 & 61.2 & 54.5 & 45.9 & 0.0 & 0.0 & 0.0\\
    & MACER & 1.183 & 83.3 & 80.1 & 75.9 & 70.4 & 64.2 & 56.7 & 47.7 & 36.0 & 0.0 & 0.0 & 0.0\\
    & Consistency & 1.212 & 84.9 & 81.1 & 76.4 & 71.2 & 65.2 & 57.8 & 49.3 & 39.2 & 0.0 & 0.0 & 0.0\\ 
    & SmoothMix & 1.237 & 84.4 & 80.7 & 76.3 & 71.2 & 65.6 & 58.9 & 52.4 & 44.2 & 0.0 & 0.0 & 0.0 \\
    \cmidrule(l){2-2} \cmidrule(l){3-3} \cmidrule(l){4-14}
    & \textbf{CAT-RS (Ours)} & \underline{\textbf{1.274}} & 82.5 & 79.6 & \textbf{76.2} & \underline{\textbf{72.4}} & \underline{\textbf{67.8}} & \underline{\textbf{62.5}} & \underline{\textbf{56.7}} & \underline{\textbf{49.0}} & 0.0 & 0.0 & 0.0\\
    \midrule
    
    \multirow{6.5}{*}{1.00}& Gaussian & 1.316 & \underline{79.0} & {74.3} & 68.6 & 62.5 & 56.2 & 50.0 & 43.1 & 36.4 & 29.2 & 23.1 & 17.5 \\
    & Stability & 1.394 & 78.1 & \underline{74.4} & \underline{70.2} & 65.5 & 59.4 & 53.3 & 46.4 & 39.9 & 32.8 & 26.2 & 19.6 \\
    & SmoothAdv & 1.538 & 77.0 & 73.7 & 69.6 & \underline{65.5} & \underline{61.3} & 56.3 & 50.9 & 45.5 & 39.1 & 32.6 & 26.9\\
    & MACER & 1.504 & 74.1 & 71.2 & 67.6 & 63.9 & 60.2 & 55.7 & 50.6 & 45.5 & 39.5 & 33.4 & 27.4\\
    & Consistency & 1.491 & 75.5 & 72.4 & 68.4 & 64.5 & 59.8 & 54.8 & 49.4 & 44.0 & 37.9 & 31.7 & 25.7 \\ 
    & SmoothMix & 1.534 & 76.4 & 72.6 & 68.3 & 63.3 & 58.4 & 53.7 & 48.6 & 43.4 & 38.4 & 33.3 & 28.3 \\
    \cmidrule(l){2-2} \cmidrule(l){3-3} \cmidrule(l){4-14}
    & \textbf{CAT-RS (Ours)} & \underline{\textbf{1.607}} & 73.8 & 71.1 & 68.0 & \textbf{64.9} & \textbf{61.1} & \underline{\textbf{57.3}} & \underline{\textbf{52.9}} & \underline{\textbf{48.0}} & \underline{\textbf{43.2}} & \underline{\textbf{37.4}} & \underline{\textbf{31.7}}\\
    \bottomrule
\end{tabular}
\caption{Comparison of ACR and approximate certified test accuracy (\%) on Fashion-MNIST. For each column, we set our result bold-faced if it improves the Gaussian baseline. We set the result underlined if it achieves the highest among the baselines.}
\label{tab:FMNIST}
\end{table*}

\subsection{Additional result on CIFAR-10}
\label{ap:curves}

\begin{figure*}[th]
\centering
\begin{adjustbox}{width=\linewidth}
 \subfigure[$\sigma=0.25$]
 {
    \includegraphics[width=0.3\linewidth]{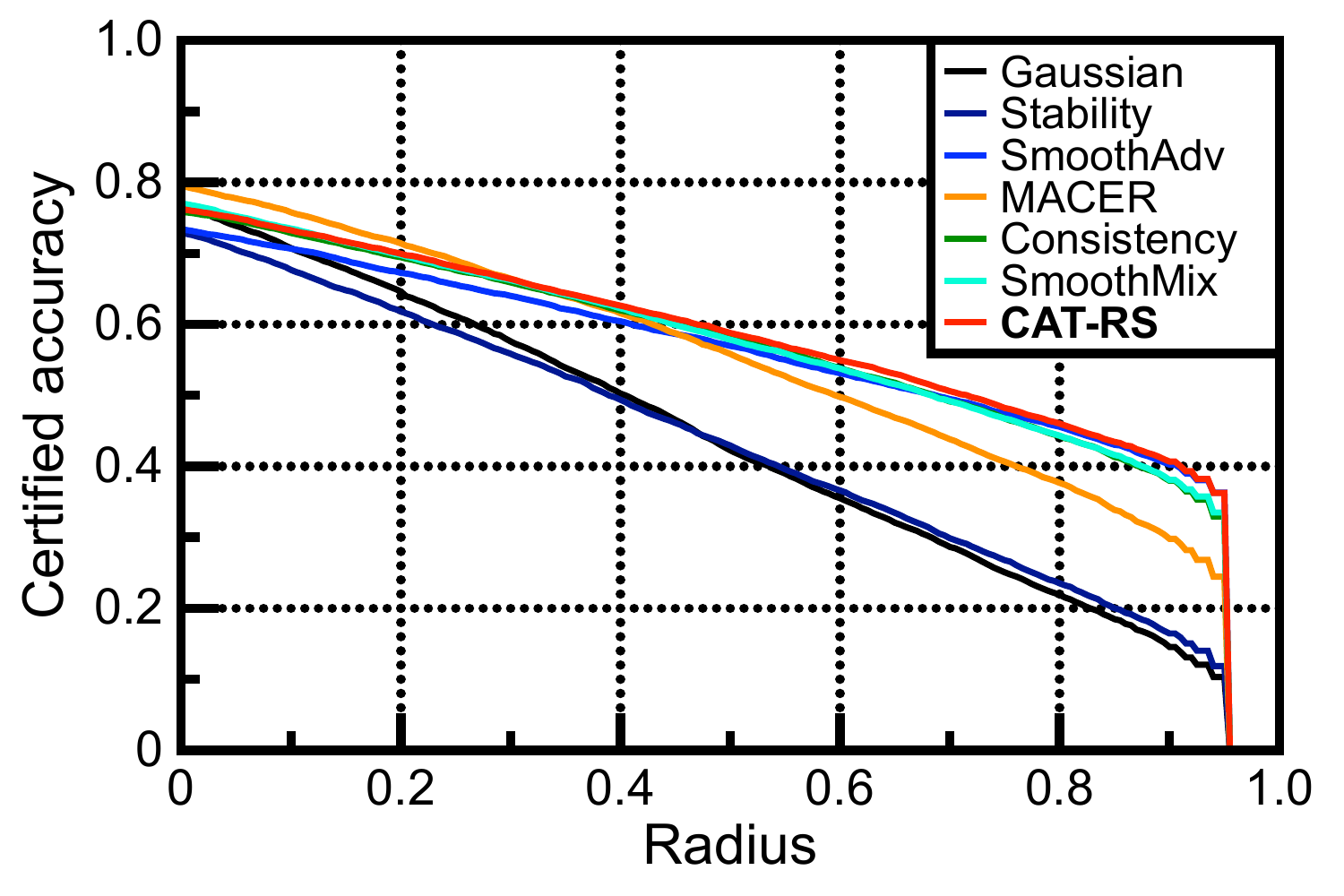}
 	\label{fig:cifar10_25}
 }
 \subfigure[$\sigma=0.50$]
 {
    \includegraphics[width=0.3\linewidth]{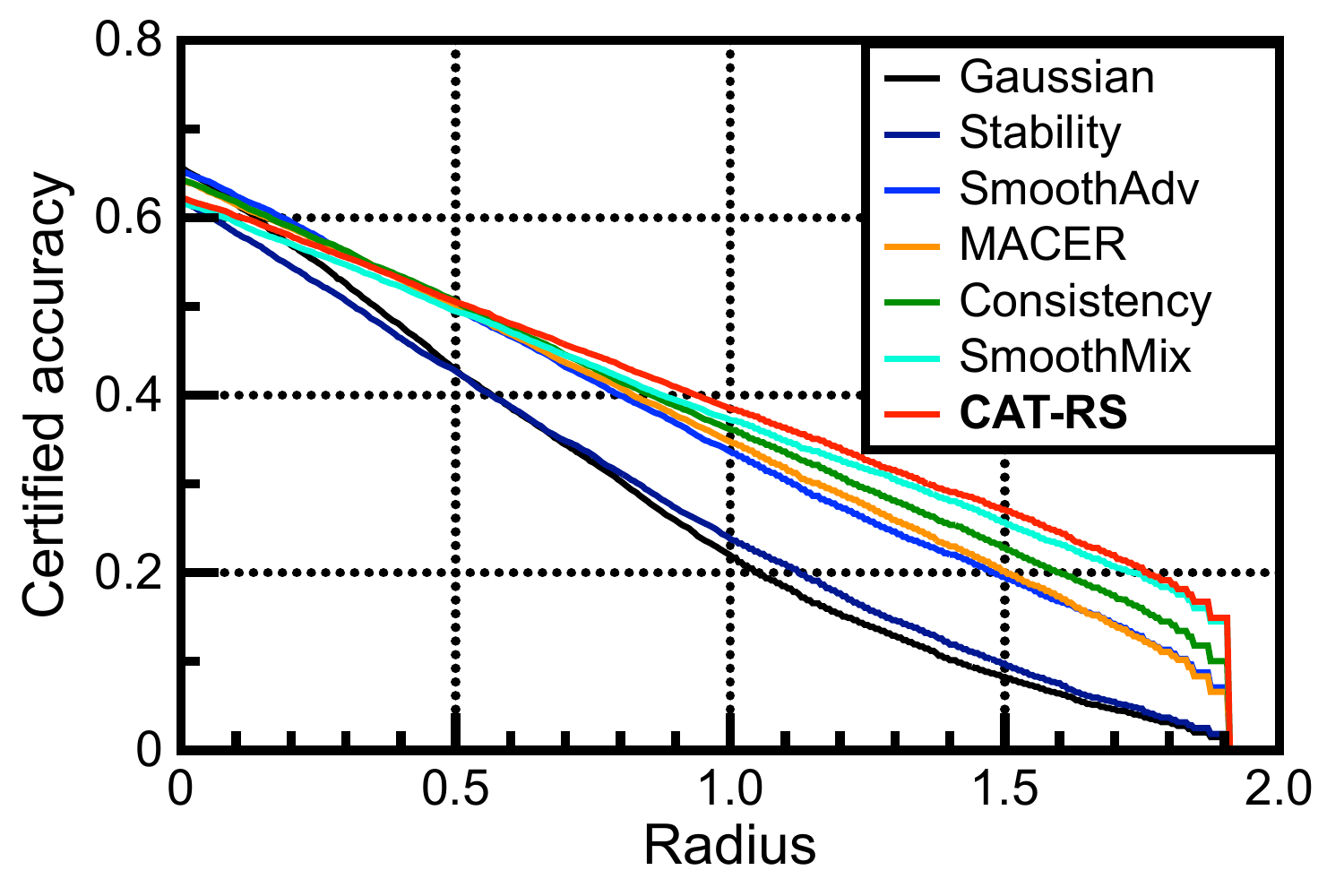}
 	\label{fig:cifar10_50}
 }
 \subfigure[$\sigma=1.00$]
 {
    \includegraphics[width=0.3\linewidth]{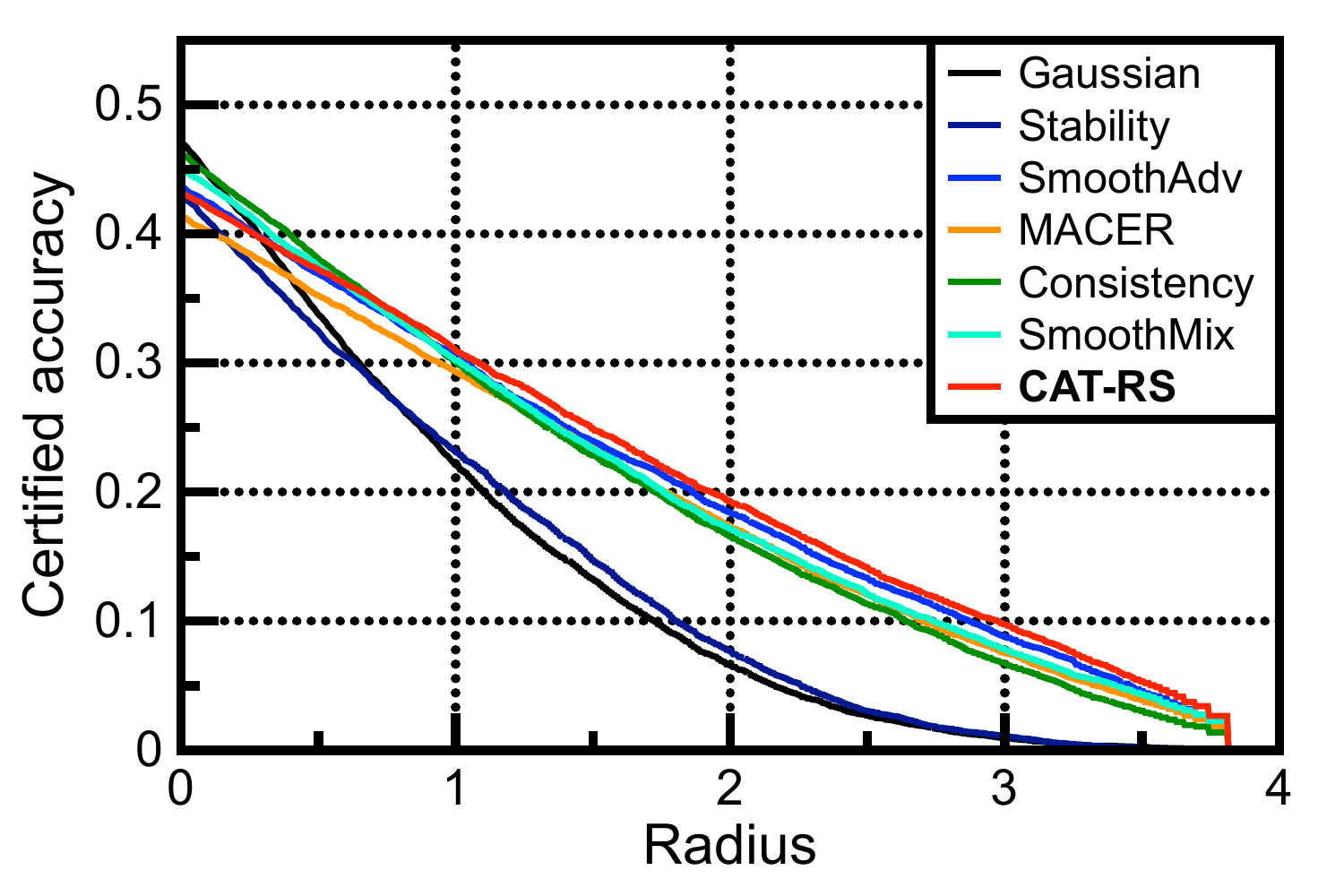}
 	\label{fig:cifar10_100}
 }
 \end{adjustbox}
 \caption{Comparison of approximate certified accuracy for various training methods on CIFAR-10. The sharp drop of certified accuracy in each plot is due to an upper bound in radius that \textsc{Certify} can output for a given $\sigma$, $N=100,000$, and $\alpha=0.001$.}
 \label{fig:cifar10}
\end{figure*}

We provide additional results on CIFAR-10 in this section. We present in Figure~\ref{fig:cifar10} the plots of the approximate certified accuracy across varying $r$. Overall, CAT-RS offers the best robustness while maintaining comparable clean accuracy. We also compare approximate certified test accuracy under $\ell_\infty$ adversary in Table~\ref{tab:l_infty_robustness}. The comparison is based on the models trained with $\sigma=0.25$, and CAT-RS achieves the highest robust accuracy. Although we mainly focus on $\ell_2$-robustness as randomized smoothing is known as the state-of-the-art on certifying against $\ell_2$ adversary, the smoothed classifiers obtained from CAT-RS can certify other adversaries with different certification methods \cite{pmlr-v119-yang20c,kumar2020curse}.

\begin{table*}[ht]
\centering
\begin{tabular}{l|cccccc|c}
    \toprule
    CIFAR-10 ($\ell_\infty$) & Gaussian & Stability & SmoothAdv & MACER & Consistency & SmoothMix & \textbf{CAT-RS} \\ 
    \midrule
    Clean ($\varepsilon=0$) & {76.6} & 73.0 & 73.4 & 79.5 & 75.8 & 77.1 & 76.3 \\
    Robust ($\varepsilon=\frac{2}{255}$) & 47.8 & 47.0 & 59.1 & 59.7 & 60.7 & 60.7 & 61.4 \\
    \bottomrule
\end{tabular}
\caption{{Comparison of $\ell_\infty$ certified accuracy (\%) on CIFAR-10 with radius $\varepsilon$. We assume $\sigma=0.25$ in this experiment.}}
\label{tab:l_infty_robustness}
\end{table*}

\subsection{Result on CIFAR-100}
\label{ap:CIFAR100_datasets}

Table~\ref{tab:cifar100} shows the results for $\sigma \in \{0.25, 0.50\}$\footnote{We omit the results for $\sigma=1.0$ as all methods achieve low clean accuracy of $\sim20\%$, which is less meaningful.} on CIFAR-100 \cite{dataset/cifar} dataset. Still, CAT-RS achieves the best ACR by boosting the robustness of the smoothed classifier. Especially, CAT-RS improves the certified accuracy over the whole range of radii while keeping the certified accuracy at $r=0.00$ comparable. For example, compared to SmoothMix for $\sigma=0.50$, CAT-RS achieves higher accuracy at $r=0.00$ by $34.0\% \rightarrow 35.4\%$ as well as at $r=1.75$ by $8.2\% \rightarrow 9.0\%$, resulting in the ACR improvement by $0.352 \rightarrow 0.372$. This result suggests that our confidence-aware training effectively plays its role.

\begin{table*}[th]
\centering
\begin{tabular}{clc|cccccccc}
    \toprule
    $\sigma$ &  Methods & ACR & 0.00 & 0.25 & 0.50 & 0.75 & 1.00 & 1.25 & 1.50 & 1.75 \\ 
    \midrule
    \multirow{6.5}{*}{0.25}& Gaussian & 0.228 & 48.9 & 33.7 & {20.9} & {12.0} & 0.0 & 0.0 & 0.0 & 0.0 \\
    & Stability & 0.159 & 34.3 & 23.4 & 14.5 & 7.8 & 0.0 & 0.0 & 0.0 & 0.0 \\ 
    & SmoothAdv & 0.298 &46.4 & 38.3 & 30.4 & 23.0 & 0.0 & 0.0 & 0.0 & 0.0 \\
    & MACER & 0.283 & \underline{51.1} & 39.5 & 28.1 & 18.1 & 0.0 & 0.0 & 0.0 & 0.0 \\  
    & Consistency & 0.263 & 39.3 & 33.1 & 26.9 & 21.0 & 0.0 & 0.0 & 0.0 & 0.0 \\
    & SmoothMix & 0.295 & 49.9 & 39.5 & 29.5 & 20.8 & 0.0 & 0.0 & 0.0 & 0.0 \\
    \cmidrule(l){2-2} \cmidrule(l){3-3} \cmidrule(l){4-11}
    & \textbf{CAT-RS (Ours)} & \underline{\textbf{0.312}} & 48.2 & \underline{\textbf{39.8}} & \underline{\textbf{31.7}} & \underline{\textbf{24.4}} & 0.0 & 0.0 & 0.0 & 0.0\\
    \midrule
    \multirow{6.5}{*}{0.50}& Gaussian & 0.259 & {36.5} & 27.8 & 20.4 & {14.7} & {10.1} & {6.8} & {4.2} & {2.3} \\
    & Stability & 0.078 & 8.6 & 7.2 & 5.9 & 4.6 & 3.7 & 2.6 & 1.9 & 1.2 \\
    & SmoothAdv & 0.342 & 36.7 & 30.5 & 24.9 & 19.9 & 15.8 & 12.0 & 9.1 & 6.3\\
    & MACER & 0.314 & \underline{37.8} & 29.7 & 23.4 & 18.2 & 14.0 & 10.3 & 7.3 & 4.7\\
    & Consistency & 0.275 & 24.3 & 21.4 & 18.5 & 16.1 & 13.8 & 11.7 & 9.3 & 7.0 \\ 
    & SmoothMix & 0.352 & 34.0 & 29.1 & 24.6 & 20.3 & 16.9 & 13.9 & 11.0 & 8.2 \\
    \cmidrule(l){2-2} \cmidrule(l){3-3} \cmidrule(l){4-11}
    & \textbf{CAT-RS (Ours)} & \underline{\textbf{0.368}} & 35.8 & \underline{\textbf{30.5}} & \underline{\textbf{25.7}} & \underline{\textbf{21.2}} & \underline{\textbf{17.5}} & \underline{\textbf{14.4}} & \underline{\textbf{11.5}} & \underline{\textbf{8.6}}\\
    \bottomrule
\end{tabular}
\caption{Comparison of ACR and approximate certified test accuracy (\%) on CIFAR-100. For each column, we set our result bold-faced when it improves the Gaussian baseline. We set our result underlined if it achieves the highest among the baselines.}
\label{tab:cifar100}
\end{table*}

\clearpage
\section{Analysis on variance of results}
\label{ap:variance}

In our experiments, we compare single-seed results of ACR and approximate certified accuracy following the evaluation protocol of the reported baselines given prior observations that ACR is quite robust to multiple runs \cite{nips_salman19,Zhai2020MACER,jeong2020consistency,jeong2021smoothmix}. Nevertheless, we further report in Table~\ref{tab:variance} a variance analysis of the reported results across 5 different random seeds.\footnote{For the CIFAR-10 experiments in Table~\ref{tab:variance}, we use the uniformly subsampled CIFAR-10 test set of size 2000, instead of the full test set: there can be discrepancy from the value reported in Table~\ref{tab:cifar10} based on the full test set.} The results indeed show that our major performance metric of ACR achieves quite robust performance over multiple runs, confirming the statistical significance of our improvements.

\begin{table*}[ht]
\centering
\begin{tabular}{l|ccc|c}
    \toprule
    Dataset & &{MNIST}& & {CIFAR-10} \\
    \midrule
    ACR & $\sigma=0.25$ & $\sigma=0.5$ & $\sigma=1.0$  & $\sigma=0.5$\\ 
    \midrule
    Gaussian & $0.9109$\stdv{0.0003} & $1.5581$\stdv{0.0016}& $1.6184$\stdv{0.0021} & $0.5406$\stdv{0.0109}\\  
    Stability & $0.9152$\stdv{0.0007} & $1.5719$\stdv{0.0028} & $1.6341$\stdv{0.0018}  & $0.5254$\stdv{0.0209}\\
    SmoothAdv & $0.9322$\stdv{0.0005} & $1.6872$\stdv{0.0007} & $1.7786$\stdv{0.0017} & $0.7009$\stdv{0.0145}\\
    MACER & $0.9201$\stdv{0.0006} & $1.5899$\stdv{0.0069} & $1.5950$\stdv{0.0051} & $0.6698$\stdv{0.0045}\\
    Consistency & $0.9279$\stdv{0.0003} & $1.6549$\stdv{0.0011} & $1.7376$\stdv{0.0017} & $0.7170$\stdv{0.0034}\\
    SmoothMix & $0.9317$\stdv{0.0002} & $1.6932$\stdv{0.0007} & $1.8185$\stdv{0.0016} & $0.7362$\stdv{0.0063}\\
    \midrule
    \textbf{CAT-RS (Ours)} & {$\mathbf{0.9329}$\stdv{0.0001}} & $\mathbf{1.7004}$\stdv{0.0005} & $\mathbf{1.8282}$\stdv{0.0018} & $\mathbf{0.7525}$\stdv{0.0028}\\
    \bottomrule
\end{tabular}
\caption{Comparison of the mean and standard deviation of ACR on MNIST and CIFAR-10. The results are calculated over 5 runs with different seeds. For each column, we set our result bold-faced if it achieves the highest ACR among the baselines.}
\label{tab:variance}
\end{table*}

\section{Analysis on the training cost}
\label{ap:training_cost}

Table~\ref{tab:training_cost} compares the training times of different methods on CIFAR-10 and their resulting ACRs. As mentioned in Section~\ref{ss:overall}, it shows that CAT-RS takes as much time as SmoothAdv and less time than SmoothMix under the same $M=4$, while achieving a better ACR. Compared to Consistency $(M=2)$, on the other hand, CAT-RS $(M=4)$ roughly takes $~2.9$ times training time: besides of the $2$ times overhead from larger $M$, it takes an extra cost from an adversarial search which is also applied for SmoothAdv and SmoothMix.

\begin{table*}[ht]
\centering
\begin{tabular}{c|cccccc}
    \toprule
    Methods & Gaussian &Consistency&SmoothAdv &SmoothMix&SmoothMix&CAT-RS (Ours) \\
    \midrule
    Number of noises ($M$) & 1 & 2 & 4 & 2 & 4 & 4\\
    Training cost (hrs) &4.6 &8.7  & 23.1 & 12.5 & 33.3 & 25.3\\
    ACR ($\sigma=0.25$) & 0.424 & 0.552 & 0.544 &0.553 & 0.558& \textbf{0.562}\\  
    \bottomrule
\end{tabular}
\caption{Comparison of the training cost and ACR on CIFAR-10. The training costs are calculated based on the GPU hours with a single NVIDIA TITAN X Pascal GPU.}
\label{tab:training_cost}
\end{table*}

\clearpage
\section{Comparison of accuracy-robustness trade-off}
\label{ap:trade-off}

\begin{figure*}[ht]
	\centering
	\subfigure[ACR]
	{
	    \includegraphics[width=0.3\linewidth]{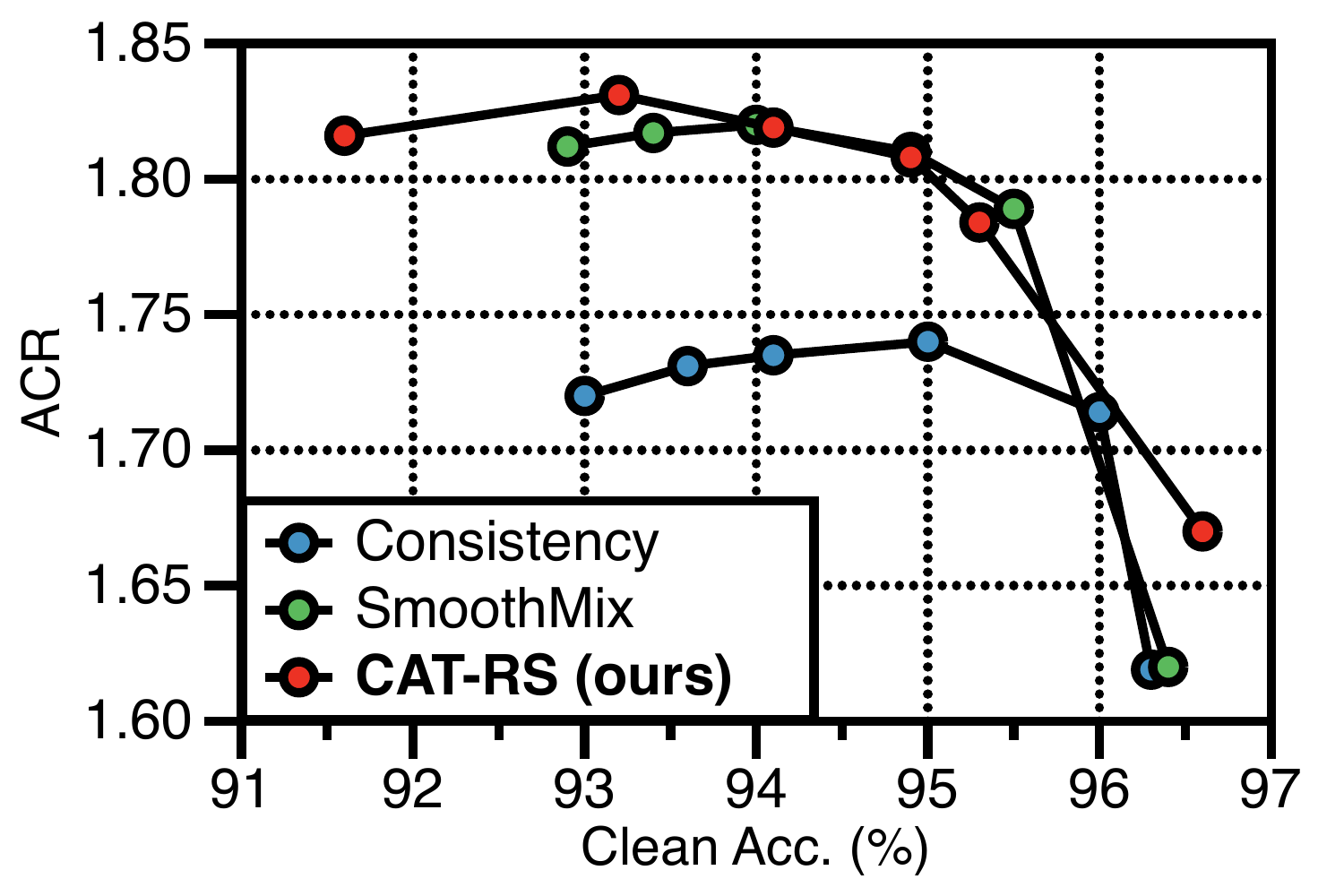}
		\label{fig:clean_acr}
	}
	\subfigure[Certified Acc., $r=1.0$]
	{
	    \includegraphics[width=0.3\linewidth]{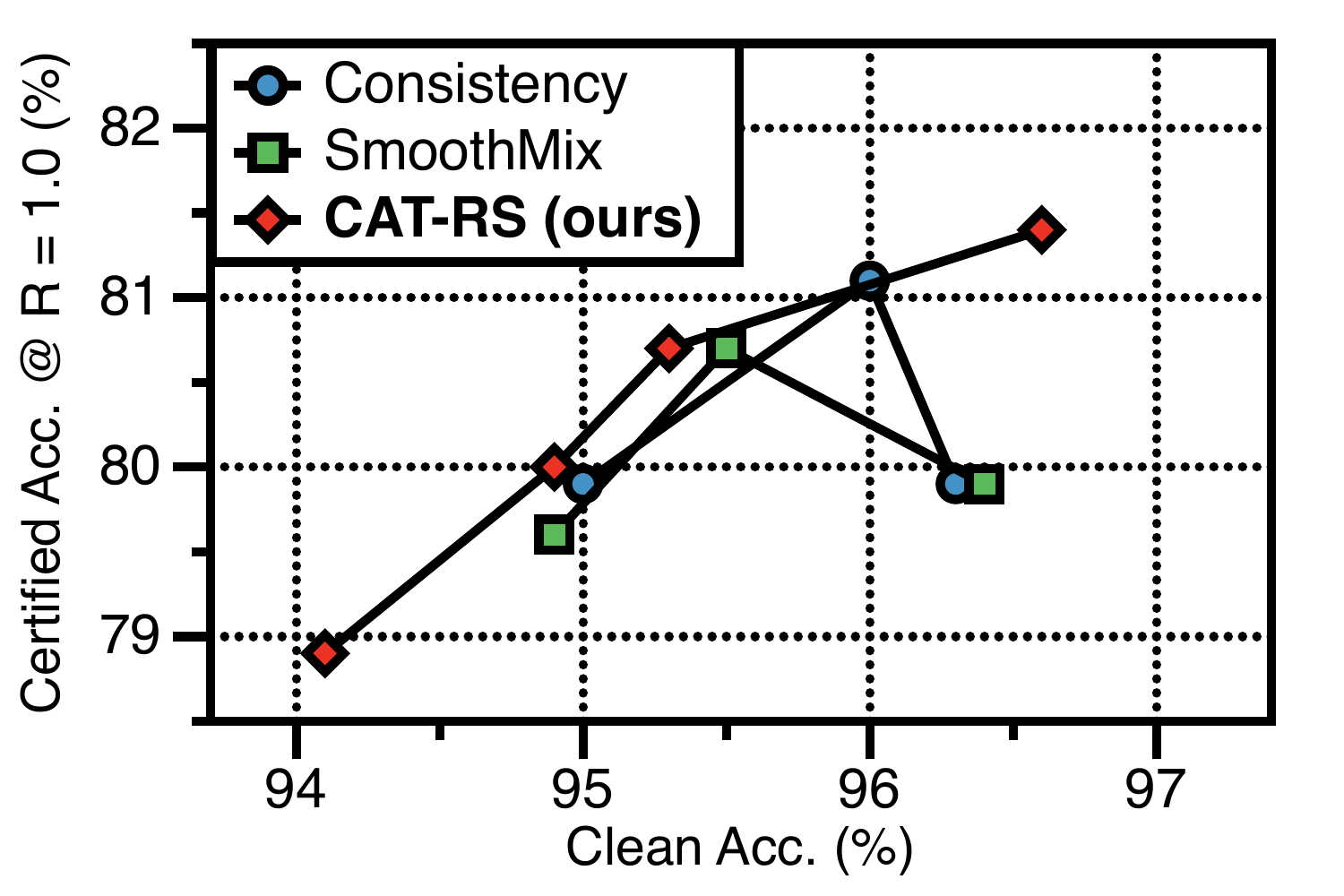}
		\label{fig:clean_r1}
	}
	\subfigure[Certified Acc., $r=2.0$]
	{
	    \includegraphics[width=0.3\linewidth]{assets/mnist_100_r2.pdf}
		\label{fig:clean_r2}
	}
	\caption{Comparison of the trends between the clean accuracy \emph{vs.} (a) ACR, (b) the certified accuracy at $r=1.0$, and (c) at $r=2.0$, that each method exhibits as varying its hyperparameter. We assume MNIST dataset with $\sigma=1.0$ for this experiment.}
	\label{fig:tradeoff_detail}
\end{figure*}

\begin{table*}[ht]
\centering
\small
\begin{adjustbox}{width=0.6\linewidth}
\begin{tabular}{ccc|cccccc}
    \toprule
    Methods & Setups & ACR & 0.00 & 0.50 & 1.00 & 1.50 & 2.00 & 2.50 \\ 
    \midrule
    Gaussian & - & 1.620 & 96.4 & 91.4 & 79.9 & 59.6 & 32.6 & 10.8  \\
    \midrule
    \multirow{5.0}{*}{Consistency}
    & $\lambda=\pz1$ & {1.714} & 96.0 & 91.2 & {81.1} & {63.5} & {39.2} & {16.2} \\
    & $\lambda=\pz5$ & {1.740} & 95.0 & 89.7 & {79.9} & {63.7} & {41.9} & {20.0} \\
    & $\lambda=10$ & {1.735} & 94.1 & 88.6 & 78.5 & {62.8} & {42.4} & {22.1}  \\
    & $\lambda=15$ & {1.731} & 93.6 & 87.7 & 77.8 & {62.3} & {42.6} & {22.9}  \\
    & $\lambda=20$ & {1.720} & 93.0 & 86.6 & 77.1 & {61.6} & {42.1} & {23.4} \\
    & $\lambda=25$ & 1.226 & 73.2 & 64.4 & 53.9 & 42.4 & 27.4 & {14.5} \\
    \midrule
    \multirow{5.0}{*}{SmoothMix}
    & $\eta=\pz1$ & {1.789} & 95.5 & 90.5 & {80.7} & {64.1} & {43.1} & {24.1} \\
    & $\eta=\pz2$ & {1.810} & 94.9 & 89.7 & 79.6 & {63.8} & {44.4} & {26.6} \\
    & $\eta=\pz4$ & {1.820} & 94.0 & 88.4 & 78.3 & {63.0} & {44.9} & {28.7} \\
    & $\eta=\pz8$ & {1.817} & 93.4 & 87.5 & 77.3 & {62.4} & {44.8} & {29.3} \\
    & $\eta=16$ & {1.812} & 92.9 & 86.7 & 76.6 & {61.8} & {44.5} & {29.6} \\
    \midrule
    \multirow{7.0}{*}{\textbf{\shortstack{CAT-RS \\ (Ours)}}}
    & $\lambda=0.00$ & {1.670} & 96.6 & 91.8 & 81.4 & 62.4 & 35.7 & 12.2 \\
    & $\lambda=0.12$ & {1.784} & 95.3 & 90.2 & 80.7 & 64.7 & 43.8 & 23.4 \\

    & $\lambda=0.25$ & {1.808} & 94.9 & 89.6 & 80.0 & 64.9 & 45.3 & 26.0 \\
    & $\lambda=0.50$ & {1.819} & 94.1 & 88.4 & 78.9 & 64.6 & 46.2 & 28.1 \\
    & $\lambda=1.00$ & {1.831} & 93.2 & 87.2 & 77.6 & 64.0 & 47.2 & 30.0 \\
    & $\lambda=2.00$ & {1.816} & 91.6 & 85.0 & 75.7 & 62.9 & 48.0 & 31.5 \\
    & $\lambda=4.00$ & {1.777} & 87.2 & 80.1 & 71.6 & 61.7 & 48.4 & 33.4 \\
    \bottomrule
\end{tabular}
\end{adjustbox}
\caption{Comparison of ACR and approximate certified test accuracy on MNIST for varying hyperparameters of three different methods: Consistency, SmoothMix, and CAT-RS (ours). We assume $\sigma=1.0$ in this experiment. ``Gaussian'' indicates the baseline Gaussian training. Consistency and SmoothMix degenerates to Gaussian when their hyperparameter is set to 0.}
\label{tab:trends}
\end{table*}

\section{Additional ablation study}
\label{ablation:detail}

\subsection{Ablation study on loss design}
\label{ap:abla_loss}

Our loss design of $L^{\mathtt{CAT}\text{-}\mathtt{RS}}$ in \eqref{eq:overall} combines several important ideas as proposed in Section~\ref{s:method}, and here we validate that each of the components has an individual effect in improving the certified robustness. In Table~\ref{tab:cifar10_ablation_method}, we compare several variants of $L^{\mathtt{CAT}\text{-}\mathtt{RS}}$, including the followings: (a) training with $L^{\mathtt{low}}$ \eqref{eq:bottom_k} only, (b) $L^{\mathtt{high}}$ \eqref{eq:worst} only, (c) $L^{\tt base}$ + $\lambda \cdot L^{\tt high}$, where $L^{\tt base}:= \frac{1}{M} \sum_{i=1}^{M} \mathbb{CE}(F(x+\delta_i), y)$ denotes the standard Gaussian training, and (d) $L^{\mathtt{low}} + \lambda \cdot L^{\mathtt{high}}$. Here, notie that (c) and (d) does not apply the masking condition $\mathds{1}[K=M]$ to $L^{\tt high}$ (Section~\ref{ss:overall}) compared to $L^{\mathtt{CAT}\text{-}\mathtt{RS}}$. 

Overall, we observe that (a) even though ACR of $L^{\mathtt{low}}$ is slightly degraded compared to $L^{\mathtt{base}}$, $L^{\mathtt{low}}$ can achive a better clean accuracy instead, and (b) when combined with $L^{\mathtt{high}}$, $L^{\mathtt{low}}$ achieves a better ACR than $L^{\mathtt{base}} + \lambda \cdot L^{\mathtt{high}}$ from a better balancing between accuracy and robustness; and (c) yet, CAT-RS further improves ACR by applying the masking strategy to $L^{\mathtt{high}}$.

Table~\ref{tab:cifar10_ablation_kl} considers three variants of $L^{\mathtt{high}}$ \eqref{eq:worst}: (a) the outer maximization \eqref{eq:worst} is replaced by averaging; (b) the label assignment $\hat{y}$ is set by $\hat{F}(x):=\frac{1}{M}\sum_{i=1}^M F(x+\delta_i)$, \ie the averaged prediction over $M$ noise samples; and (c) the label assignment $\hat{y}$ is set by the hard label $y$. The results show that our form of worst-case loss achieves the best performance in terms of ACR, confirming that both designs of (a) maximizing loss over noise samples, and (b) utilizing soft-labeled $\hat{y}$'s in $L^{\mathtt{high}}$ work effectively.

\begin{table*}[ht]
\centering
\begin{adjustbox}{width=0.75\linewidth}
\begin{tabular}{l|ccc|c|cccccccccc}
    \toprule
    Method (CIFAR-10) & $L^{\tt low}$ & $L^{\tt High}$ & Mask & ACR & 0.00 & 0.25 & 0.50 & 0.75 & 1.00 & 1.25 & 1.50 & 1.75 \\ 
    \midrule
    $L^{\tt base}$ (Gaussian; \eqref{eq:gaussian_training}) & $L^{\tt base}$ & \xmark & - & 0.523 & 66.2 & 55.2 & 42.9 & 31.0 & 21.3 & 14.4 & 7.9 & 3.7 \\
    \cmidrule{1-1} \cmidrule(l){2-4} \cmidrule(l){5-13}
    (a) $L^{\tt low}$ only & \cmark & \xmark & - & 0.508 & 67.0 & 54.6 & 41.9 & 29.7 & 20.4 & 13.1 & 7.6 & 3.6 \\
    (b) $L^{\tt high}$ only & \xmark & \cmark & \xmark & 0.685 & 55.2 & 48.7 & 44.0 & 39.9 & 34.8 & 30.7 & 26.5 & 20.7 \\
    (c) $L^{\tt base} + \lambda \cdot L^{\tt high}$ & $L^{\tt base}$ & \cmark & \xmark & 0.694 & 62.4 & 54.4 & 48.1 & 41.4 & 34.4 & 28.1 & 22.5 & 17.6\\
    (d) $L^{\tt low} + \lambda \cdot L^{\tt high}$ & \cmark & \cmark & \xmark & 0.706 & 59.7 & 54.6 & 48.2 & 41.2 & 35.5 & 30.1 & 23.6 & 18.5 \\
    \cmidrule{1-1} \cmidrule(l){2-4} \cmidrule(l){5-13}
    \textbf{$L^{\mathtt{CAT}\text{-}\mathtt{RS}}$ (Ours; \eqref{eq:overall})} & \cmark & \cmark & \cmark & {0.710} & 57.7 & 52.7 & 48.4 & 41.6 & 36.2 & 29.7 & 25.3 & 20.6 \\
    \bottomrule
\end{tabular}
\end{adjustbox}
\caption{Comparison of ACR and certified accuracy (\%) for ablations of CAT-RS. All the models are on CIFAR-10 with $\sigma=0.5$. $L^{\tt base}$ as mark indicates the use of Gaussian training \eqref{eq:gaussian_training}. We mark ``Mask'' if we apply $\mathds{1}[K=M]$ to $L^{\tt high}$ in \eqref{eq:overall}.}
\label{tab:cifar10_ablation_method}
\vspace{0.05in}
\begin{adjustbox}{width=0.75\linewidth}
\begin{tabular}{lc|cccccccccc}
    \toprule
    Method (CIFAR-10) & ACR & 0.00 & 0.25 & 0.50 & 0.75 & 1.00 & 1.25 & 1.50 & 1.75 \\ 
    \midrule
    (a) $\frac{1}{M}\sum_i \left(\max_{\delta^*_i}  \mathrm{KL}(F(x+\delta^*_i), \hat{y})\right)$ & 0.694 & 61.2 & 53.5 & 46.7 & 41.0 & 34.1 & 29.3 & 23.6 & 18.2\\
    (b) $\max_{i, \delta^*_i}  \mathrm{KL}(F(x+\delta^*_i), \hat{F}(x))$ & 0.694 & 57.2 & 51.8 & 46.9 & 40.7 & 34.7 & 30.7 & 24.4 & 18.7 \\
    (c) {$\max_{i, \delta^*_i}  \mathrm{KL}(F(x+\delta^*_i), y)$} & 0.701 & 56.4 & 51.5 & 46.3 & 39.8 & 36.0 & 30.6 & 25.8 & 20.9\\
    \cmidrule{1-1} \cmidrule(l){2-2} \cmidrule(l){3-10}
    \textbf{$\max_{i, \delta^*_i}  \mathrm{KL}(F(x+\delta^*_i), \hat{y})$ ($L^{\tt high}$; Ours)} & {0.710} & 57.7 & 52.7 & 48.4 & 41.6 & 36.2 & 29.7 & 25.3 & 20.6 \\
    \bottomrule
\end{tabular}
\end{adjustbox}
\caption{Comparison of ACR and certified accuracy (\%) ablations of $L^{\tt high}$ \eqref{eq:worst}. All the models are on CIFAR-10 with $\sigma=0.5$.}
\label{tab:cifar10_ablation_kl}
\end{table*}

\subsection{Detailed results on ablation study}
\label{ap:abla}

\begin{table*}[ht]
\begin{minipage}{.49\textwidth}
\centering
\begin{adjustbox}{width=\linewidth}
\begin{tabular}{cc|cccccccc}
    \toprule
    \multicolumn{2}{c}{CIFAR-10} & \multicolumn{8}{c}{Certified accuracy (\%)} \\ 
    \cmidrule{1-2} \cmidrule(l){3-10}
     Setups & ACR & 0.0 & 0.25 & 0.5 & 0.75 & 1.0 & 1.25 & 1.5 & 1.75 \\ 
    \midrule
    $\lambda=0.25$ & 0.684 & 63.4 & 55.6 & 48.1 & 40.4 & 33.6 & 27.1 & 21.2 & 15.2\\

    $\lambda=0.50$ & 0.692 & 60.9 & 54.1 & 47.6 & 40.2 & 35.0 & 27.9 & 23.5 & 18.2\\
    $\lambda=1.00$ & 0.710 & 57.7 & 52.7 & 48.4 & 41.6 & 36.2 & 29.7 & 25.3 & 20.6\\
    $\lambda=2.00$ & 0.703 & 54.2 & 50.3 & 45.2 & 39.9 & 35.5 & 31.9 & 27.8 & 22.1\\
    $\lambda=4.00$ & 0.698 & 52.6 & 48.6 & 44.2 & 39.7 & 36.6 & 32.7 & 27.2 & 22.9\\
    
    \bottomrule
\end{tabular}
\end{adjustbox}
\caption{Comparison of ACR and approximate certified test accuracy (\%) for varying $\lambda$ on CIFAR-10. We assume $\sigma=0.5$.}
\label{tab:cifar10_ablation_lbd}
\end{minipage}
\hspace*{\fill}
\begin{minipage}{.47\textwidth}
\centering

\begin{adjustbox}{width=\linewidth}
\begin{tabular}{cc|cccccccc}
    \toprule
    \multicolumn{2}{c}{CIFAR-10} & \multicolumn{8}{c}{Certified accuracy (\%)} \\ 
    \cmidrule{1-2} \cmidrule(l){3-10}
    
     Setups & ACR  & 0.0 & 0.25 & 0.5 & 0.75 & 1.0 & 1.25 & 1.5 & 1.75 \\ 
    \midrule
    $M=1$ & 0.661 & 66.2 & 55.2 & 42.9 & 31.0 & 21.3 & 14.4 & 7.9 & 3.7\\

    $M=2$ & 0.684 &61.2 & 54.2 & 47.5 & 40.5 & 32.8 & 28.1 & 21.9 & 17.4\\
    $M=4$ & {0.710} & 57.7 & 52.7 & 48.4 & 41.6 & 36.2 & 29.7 & 25.3 & 20.6\\
    $M=8$ & 0.697 & 54.7 & 50.2 & 45.0 & 40.1 & 36.4 & 31.3 & 25.9 & 21.6\\
    
    \bottomrule
\end{tabular}
\end{adjustbox}
\caption{Comparison of ACR and approximative certified test accuracy (\%) for varying $M$ on CIFAR-10. We assume $\sigma=0.5$.}
\label{tab:cifar10_abla_m}
\end{minipage}
\hspace*{\fill}
\end{table*}

\section{Detailed results on CIFAR-10-C}
\label{ap:detail_cifar10c}

In this section, we report the detailed results on CIFAR-10-C test dataset, \ie ACR and the certified accuracy for each corruption severity and type. Our method consistently achieves the best mACR and mAcc among the baselines over severities.\footnote{The dataset is hosted at \url{https://zenodo.org/record/2535967\#.Yisixi8RpQI}.}

\begin{figure*}[ht]%
\centering
\subfigure[][Clean]{%
\label{fig:cifar10c_a}%
\includegraphics[height=0.75in]{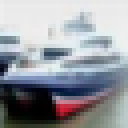}}%
\hspace{5pt}%
\subfigure[][Gaussian]{%
\label{fig:cifar10c_b}%
\includegraphics[height=0.75in]{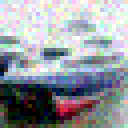}}%
\hspace{5pt}%
\subfigure[][Shot]{%
\label{fig:cifar10c_c}%
\includegraphics[height=0.75in]{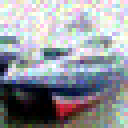}}%
\hspace{5pt}%
\subfigure[][Impulse]{%
\label{fig:cifar10c_d}%
\includegraphics[height=0.75in]{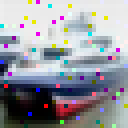}}
\subfigure[][Defocus]{%
\label{fig:cifar10c_e}%
\includegraphics[height=0.75in]{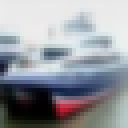}}%
\hspace{5pt}%
\subfigure[][Glass]{%
\label{fig:cifar10c_f}%
\includegraphics[height=0.75in]{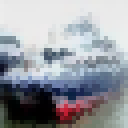}}%
\hspace{5pt}%
\subfigure[][Motion]{%
\label{fig:cifar10c_g}%
\includegraphics[height=0.75in]{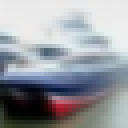}}%
\hspace{5pt}%
\subfigure[][Zoom]{%
\label{fig:cifar10c_h}%
\includegraphics[height=0.75in]{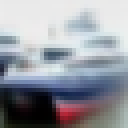}}\\
\subfigure[][Snow]{%
\label{fig:cifar10c_i}%
\includegraphics[height=0.75in]{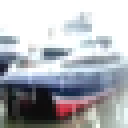}}%
\hspace{5pt}%
\subfigure[][Frost]{%
\label{fig:cifar10c_j}%
\includegraphics[height=0.75in]{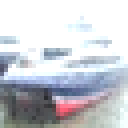}}%
\hspace{5pt}%
\subfigure[][Fog]{%
\label{fig:cifar10c_k}%
\includegraphics[height=0.75in]{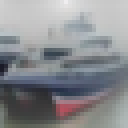}}%
\hspace{5pt}%
\subfigure[][Bright]{%
\label{fig:cifar10c_l}%
\includegraphics[height=0.75in]{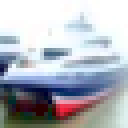}}
\subfigure[][Contrast]{%
\label{fig:cifar10c_m}%
\includegraphics[height=0.75in]{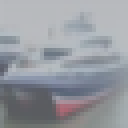}}%
\hspace{5pt}%
\subfigure[][Elastic]{%
\label{fig:cifar10c_n}%
\includegraphics[height=0.75in]{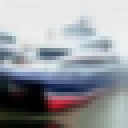}}%
\hspace{5pt}%
\subfigure[][Pixel]{%
\label{fig:cifar10c_o}%
\includegraphics[height=0.75in]{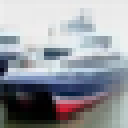}}%
\hspace{5pt}%
\subfigure[][JPEG]{%
\label{fig:cifar10c_p}%
\includegraphics[height=0.75in]{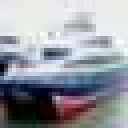}}
\caption{Images in CIFAR-10-C:
\subref{fig:cifar10c_a} is a clean test image in CIFAR-10 dataset, and the other images are the corresponding corrupted images contained in CIFAR-10-C. All corrupted images are drawn from severity 3.}
\label{fig:cifar-10-c-images}
\end{figure*}

\begin{table*}[ht]
\centering
\small
\begin{tabular}{l|cccccccccccc}
    \toprule
    
     & \multicolumn{6}{c}{Average Certified Radius}& \multicolumn{6}{c}{Certifed Test Accuracy (\%)} \\ 
    \cmidrule(l){2-7} \cmidrule(l){8-13}
    Severity & 1 & 2 & 3 & 4 & 5 & mACR & 1 & 2 & 3 & 4 & 5 & mAcc \\ 
    \midrule
    {Gaussian} & 0.392 &  0.363 &  0.342 &0.319 & 0.298 & 0.343 &68.6 &66.4 & 64.7 &  62.9  & 59.6 & 64.4 \\
    {Stability} & 0.341 &  0.319  & 0.299 &0.286 & 0.267 & 0.302 &67.0 & 63.1& 60.1 &  58.4  & 55.0 & 60.7\\
    {SmoothAdv} & \underline{0.490} & 0.465 & \underline{0.449} & \underline{0.428} & 0.404 & \underline{0.447} & 68.1 & 65.2 & 63.7 & 62.7 & 58.6 & 63.7\\ 
    {MACER} & 0.457 & 0.431 & 0.409 & 0.385 & 0.364 & 0.409 & \underline{73.5} & \underline{71.5} & \underline{69.0} & 66.4 & \underline{63.5} & \underline{68.8}\\
    {Consistency} & 0.488 & 0.463 & 0.442 & 0.424 & 0.402 & 0.444 & 69.5 & 67.1 & 65.4 & 63.9 & 62.0 & 65.6\\
    {SmoothMix} & \underline{0.490} & \underline{0.466} & 0.445 & 0.422 & \underline{0.405} & 0.446 & 72.1 & 69.5 & 66.8 & \underline{66.8} & 63.3 & 67.7\\
    \midrule 
    {\textbf{CAT-RS (Ours)}} & \textbf{0.521} & \textbf{0.493} & \textbf{0.476} & \textbf{0.458} & \textbf{0.430} & \textbf{0.475} & \textbf{75.3} &\textbf{71.6} &  \textbf{69.8} & \textbf{69.4} & \textbf{64.4} & \textbf{70.1}\\ 
    \bottomrule
\end{tabular}
\caption{Comparison of ACR and certified accuracy at $r=0.0$ on CIFAR-10-C. We report the results for five different corruption severities. For each column, we set the best and runner-up values bold-faced and underlined, respectively.}
\label{tab:cifar10_sev}
\end{table*}

\begin{table*}[ht]
\begin{minipage}{.5\textwidth}
\centering
\begin{adjustbox}{height=1.45in}
\begin{tabular}{l|cccccc|c}
    Type & \rot{Gaussian} & \rot{Stability} & \rot{SmoothAdv} & \rot{MACER} & \rot{Consistency} & \rot{SmoothMix} & \rot{\textbf{CAT-RS (Ours)}} \\ 
    \midrule
    Gaussian & 0.419 & 0.358 & 0.509 & 0.479 & 0.506 & \underline{0.511} & \textbf{0.549}\\
    Shot & 0.422 & 0.365 & 0.512 & 0.480 & 0.509 & \underline{0.514} & \textbf{0.550}\\
    Impulse & 0.417 & 0.354 & 0.507 & 0.477 & 0.507 & \underline{0.510} & \textbf{0.546}\\
    Defocus & 0.416 & 0.360 & 0.505 & 0.478 & 0.506 & \underline{0.512} & \textbf{0.544}\\
    Glass & 0.377 & 0.312 & 0.481 & 0.451 & 0.484 & \underline{0.496} & \textbf{0.512}\\
    Motion & 0.394 & 0.341 & 0.483 & 0.449 & 0.482 & \underline{0.497} & \textbf{0.517}\\
    Zoom & 0.367 & 0.329 & 0.487 & 0.442 & 0.483 & \underline{0.501} & \textbf{0.520}\\
    Snow & 0.412 & 0.362 & \underline{0.516} & 0.482 & 0.515 & 0.510 & \textbf{0.544}\\
    Frost & 0.365 & 0.359 & \underline{0.488} & 0.443 & 0.487 & 0.482 & \textbf{0.511}\\
    Fog & 0.360 & 0.310 & \underline{0.466} & 0.436 & 0.460 & 0.453 & \textbf{0.485}\\
    Bright & 0.421 & 0.375 & \underline{0.517} & 0.480 & 0.512 & 0.514 & \textbf{0.553}\\
    Contrast & 0.332 & 0.272 & \underline{0.441} & 0.403 & 0.435 & 0.424 & \textbf{0.444}\\
    Elastic & 0.337 & 0.299 & 0.421 & 0.407 & \underline{0.422} & 0.411 & \textbf{0.446}\\
    Pixel & 0.422 & 0.361 & 0.509 & 0.477 & 0.509 & \underline{0.514} & \textbf{0.548}\\
    JPEG & 0.420 & 0.361 & \underline{0.510} & 0.476 & 0.505 & 0.508 & \textbf{0.543}\\
    \midrule
    \textbf{mACR} & 0.392 & 0.341 & \underline{0.490} & 0.457 & 0.488 & \underline{0.490} & \textbf{0.521}\\
    \bottomrule
\end{tabular}
\end{adjustbox}
\caption{Comparison of \emph{average certified radius} (ACR) on CIFAR-10-C of severity 1. We set the highest values bold-faced for each row. We set the runner-up values underlined.}
\label{tab:cifar10c_sev1_rotated}
\end{minipage}
\hspace*{\fill}
\begin{minipage}{.45\textwidth}
\centering
\begin{adjustbox}{height=1.45in}
\begin{tabular}{l|cccccc|c}
    Type & \rot{Gaussian} & \rot{Stability} & \rot{SmoothAdv} & \rot{MACER} & \rot{Consistency} & \rot{SmoothMix} & \rot{\textbf{CAT-RS (Ours)}} \\ 
    \midrule
    Gaussian & 70.0 & 67.0 & 71.0 & 72.0 & 70.0 & \underline{73.0} & \textbf{77.0}\\
    Shot & 72.0 & 68.0 & 70.0 & \underline{74.0} & 71.0 & \underline{74.0} & \textbf{77.0}\\
    Impulse & 69.0 & 69.0 & 69.0 & \underline{75.0} & 71.0 & 74.0 & \textbf{78.0}\\
    Defocus & 69.0 & 68.0 & 69.0 & \underline{73.0} & 69.0 & 71.0 & \textbf{77.0}\\
    Glass & 67.0 & 65.0 & 67.0 & \underline{72.0} & 69.0 & 71.0 & \textbf{75.0}\\
    Motion & 66.0 & 66.0 & 68.0 & \textbf{74.0} & \underline{72.0} & 71.0 & \underline{72.0}\\
    Zoom & 68.0 & 67.0 & 70.0 & \underline{74.0} & 67.0 & 73.0 & \textbf{75.0}\\
    Snow & 71.0 & 68.0 & 68.0 & \underline{77.0} & 70.0 & 74.0 & \textbf{79.0}\\
    Frost & 71.0 & 66.0 & 68.0 & \textbf{76.0} & 72.0 & 72.0 & \underline{74.0}\\
    Fog & 68.0 & 67.0 & 69.0 & \underline{72.0} & 70.0 & \textbf{74.0} & \underline{72.0}\\
    Bright & 71.0 & 70.0 & 67.0 & \underline{76.0} & 71.0 & 75.0 & \textbf{80.0}\\
    Contrast & 66.0 & 62.0 & 64.0 & \textbf{72.0} & 67.0 & 69.0 & \underline{70.0}\\
    Elastic& 66.0 & 64.0 & 62.0 & \underline{69.0} & 62.0 & 65.0 & \textbf{70.0}\\
    Pixel & 67.0 & 69.0 & 69.0 & \underline{75.0} & 70.0 & 73.0 & \textbf{77.0}\\
    JPEG & 68.0 & 69.0 & 70.0 & 71.0 & 71.0 & \underline{73.0} & \textbf{77.0}\\
    \midrule 
    \textbf{mAcc} & 68.6 & 67.0 & 68.1 & \underline{73.5} & 69.5 & 72.1 & \textbf{75.3}\\
    \bottomrule
\end{tabular}
\end{adjustbox}
\caption{Comparison of certified accuracy at $r=0.0$ (\%) on CIFAR-10-C of severity 1. {We set the highest and runner-up values bold-faced and underlined, respectively.}}
\label{tab:cifar10c_sev1_clean_rotated}
\end{minipage}
\hspace*{\fill}
\end{table*}

\begin{table*}[ht]
\begin{minipage}{.5\textwidth}
\centering
\begin{adjustbox}{height=1.45in}
\begin{tabular}{l|cccccc|c}
    Type & \rot{Gaussian} & \rot{Stability} & \rot{SmoothAdv} & \rot{MACER} & \rot{Consistency} & \rot{SmoothMix} & \rot{\textbf{CAT-RS (Ours)}} \\ 
    \midrule
    Gaussian & 0.414 & 0.356 & 0.510 & 0.476 & 0.506 & \underline{0.515} & \textbf{0.546}\\
    Shot & 0.419 & 0.360 & 0.505 & 0.477 & 0.507 & \underline{0.511} & \textbf{0.544}\\
    Impulse & 0.411 & 0.345 & 0.502 & 0.467 & 0.498 & \underline{0.506} & \textbf{0.538}\\
    Defocus & 0.397 & 0.344 & 0.494 & 0.464 & 0.497 & \underline{0.506} & \textbf{0.530}\\
    Glass & 0.363 & 0.303 & 0.481 & 0.435 & 0.485 & \underline{0.497} & \textbf{0.514}\\
    Motion & 0.372 & 0.338 & 0.464 & 0.440 & 0.479 & \underline{0.493} & \textbf{0.512}\\
    Zoom & 0.361 & 0.325 & 0.477 & 0.436 & 0.474 & \underline{0.491} & \textbf{0.514}\\
    Snow & 0.361 & 0.334 & 0.470 & 0.444 & \underline{0.482} & 0.470 & \textbf{0.512}\\
    Frost & 0.321 & 0.340 & \textbf{0.475} & 0.421 & 0.444 & 0.447 & \underline{0.465}\\
    Fog & 0.251 & 0.200 & \underline{0.355} & 0.348 & 0.349 & 0.335 & \textbf{0.359}\\
    Bright & 0.413 & 0.378 & \underline{0.512} & 0.472 & 0.509 & 0.505 & \textbf{0.555}\\
    Contrast & 0.166 & 0.136 & \textbf{0.269} & 0.229 & 0.242 & 0.233 & \underline{0.253}\\
    Elastic & 0.359 & 0.307 & 0.453 & 0.420 & 0.457 & \underline{0.464} & \textbf{0.467}\\
    Pixel & 0.417 & 0.360 & 0.505 & 0.468 & 0.505 & \underline{0.513} & \textbf{0.544}\\
    JPEG & 0.415 & 0.355 & 0.500 & 0.472 & 0.504 & \underline{0.506} & \textbf{0.536}\\
    \midrule
    \textbf{mACR} & 0.363 & 0.319 &0.465 &0.431&0.463&\underline{0.466}&\textbf{0.493}\\
    \bottomrule
\end{tabular}
\end{adjustbox}
\caption{Comparison of \emph{average certified radius} (ACR) on CIFAR-10-C of severity 2. We set the highest values bold-faced for each row. We set the runner-up values underlined.}
\label{tab:cifar10c_sev2_rotated}
\end{minipage}
\hspace*{\fill}
\begin{minipage}{.45\textwidth}
\centering
\begin{adjustbox}{height=1.45in}
\begin{tabular}{l|cccccc|c}
    Type & \rot{Gaussian} & \rot{Stability} & \rot{SmoothAdv} & \rot{MACER} & \rot{Consistency} & \rot{SmoothMix} & \rot{\textbf{CAT-RS (Ours)}} \\ 
    \midrule
    Gaussian & 70.0 & 65.0 & 70.0 & 72.0 & 68.0 & 73.0 & \textbf{76.0}\\
    Shot & 70.0 & 69.0 & 68.0 & \underline{74.0} & 69.0 & 72.0 & \textbf{76.0}\\
    Impulse & 70.0 & 63.0 & 70.0 & \underline{74.0} & 71.0 & \underline{74.0} & \textbf{75.0}\\
    Defocus & 65.0 & 66.0 & 68.0 & \underline{73.0} & 69.0 & 70.0 & \textbf{76.0}\\
    Glass & 65.0 & 61.0 & 68.0 & \textbf{74.0} & 67.0 & 70.0 & \underline{72.0}\\
    Motion & 69.0 & 64.0 & 68.0 & \underline{74.0} & 73.0 & 72.0 & \textbf{75.0}\\
    Zoom & 66.0 & 66.0 & 69.0 & 72.0 & 67.0 & \underline{73.0} & \textbf{75.0}\\
    Snow & 69.0 & 66.0 & 64.0 & \underline{74.0} & 70.0 & \underline{74.0} & \textbf{76.0}\\
    Frost & 65.0 & 70.0 & 67.0 & \underline{71.0} & \underline{71.0} & \textbf{74.0} & 69.0\\
    Fog & \textbf{65.0} & 53.0 & 55.0 & \textbf{65.0} & 59.0 & \underline{60.0} & 58.0\\
    Bright & 74.0 & 69.0 & 68.0 & \underline{77.0} & 73.0 & 74.0 & \textbf{79.0}\\
    Contrast & \underline{49.0} & 32.0 & 42.0 & \textbf{50.0} & 42.0 & 44.0 & 43.0\\
    Elastic & 64.0 & 65.0 & 65.0 & \textbf{76.0} & 69.0 & 70.0 & \underline{71.0}\\
    Pixel & 67.0 & 69.0 & 68.0 & \underline{75.0} & 69.0 & 72.0 & \textbf{78.0}\\
    JPEG & 68.0 & 68.0 & 68.0 & \underline{71.0} & 69.0 & 70.0 & \textbf{75.0}\\
    \midrule 
    \textbf{mAcc} & 66.4 & 63.1 & 65.2 & \underline{71.5} & 67.1 &69.5 & \textbf{71.6}\\
    \bottomrule
\end{tabular}
\end{adjustbox}
\caption{Comparison of certified accuracy at $r=0.0$ (\%) on CIFAR-10-C of severity 2. {We set the highest and runner-up values bold-faced and underlined, respectively.}}
\label{tab:cifar10c_sev2_clean_rotated}
\end{minipage}
\hspace*{\fill}
\end{table*}

\begin{table*}[ht]
\begin{minipage}{.5\textwidth}
\centering
\begin{adjustbox}{height=1.45in}
\begin{tabular}{l|cccccc|c}
    Type & \rot{Gaussian} & \rot{Stability} & \rot{SmoothAdv} & \rot{MACER} & \rot{Consistency} & \rot{SmoothMix} & \rot{\textbf{CAT-RS (Ours)}} \\ 
    \midrule
    Gaussian & 0.414 & 0.349 & 0.504 & 0.477 & 0.506 & \underline{0.515} & \textbf{0.542}\\
    Shot & 0.410 & 0.348 & 0.505 & 0.469 & 0.500 & \underline{0.506} & \textbf{0.542}\\
    Impulse & 0.397 & 0.327 & 0.500 & 0.454 & 0.493 & \underline{0.502} & \textbf{0.528}\\
    Defocus & 0.376 & 0.330 & 0.484 & 0.447 & 0.485 & \underline{0.494} & \textbf{0.514}\\
    Glass & 0.355 & 0.301 & 0.480 & 0.433 & 0.479 & \underline{0.491} & \textbf{0.513}\\
    Motion & 0.337 & 0.302 & 0.455 & 0.410 & 0.464 & \underline{0.472} & \textbf{0.481}\\
    Zoom & 0.347 & 0.315 & 0.466 & 0.422 & 0.462 & \underline{0.478} & \textbf{0.503}\\
    Snow & 0.370 & 0.328 & 0.462 & 0.436 & \underline{0.477} & 0.458 & \textbf{0.509}\\
    Frost & 0.287 & 0.276 & \textbf{0.436} & 0.365 & 0.382 & 0.381 & \underline{0.420}\\
    Fog & 0.173 & 0.126 & \underline{0.291} & 0.249 & 0.269 & 0.253 & \textbf{0.301}\\
    Bright & 0.392 & 0.375 & \underline{0.504} & 0.459 & \underline{0.504} & 0.490 & \textbf{0.548}\\
    Contrast & 0.113 & 0.107 & \textbf{0.205} & 0.158 & 0.175 & 0.166 & \underline{0.190}\\
    Elastic & 0.338 & 0.298 & 0.436 & 0.417 & 0.435 & \underline{0.456} & \textbf{0.465}\\
    Pixel & 0.405 & 0.353 & 0.500 & 0.467 & 0.499 & \underline{0.507} & \textbf{0.537}\\
    JPEG & 0.413 & 0.351 & 0.501 & 0.473 & 0.502 & \underline{0.504} & \textbf{0.540}\\
    \midrule
    \textbf{mACR} & 0.342 & 0.299 & \underline{0.449} & 0.409 & 0.442 & 0.445 & \textbf{0.476}\\
    \bottomrule
\end{tabular}
\end{adjustbox}
\caption{Comparison of \emph{average certified radius} (ACR) on CIFAR-10-C of severity 3. We set the highest values bold-faced for each row. We set the runner-up values underlined.}
\label{tab:cifar10c_sev3_rotated}
\end{minipage}
\hspace*{\fill}
\begin{minipage}{.45\textwidth}
\centering
\begin{adjustbox}{height=1.45in}
\begin{tabular}{l|cccccc|c}
    Type & \rot{Gaussian} & \rot{Stability} & \rot{SmoothAdv} & \rot{MACER} & \rot{Consistency} & \rot{SmoothMix} & \rot{\textbf{CAT-RS (Ours)}} \\ 
    \midrule
    Gaussian & 72.0 & 66.0 & 71.0 & \underline{73.0} & 70.0 & \textbf{76.0} & \textbf{76.0}\\
    Shot & 69.0 & 64.0 & 69.0 & \underline{73.0} & 69.0 & \underline{73.0} & \textbf{76.0}\\
    Impulse & 70.0 & 60.0 & 69.0 & \underline{73.0} & 71.0 & \underline{73.0} & \textbf{74.0}\\
    Defocus & 64.0 & 66.0 & 69.0 & \underline{71.0} & 70.0 & \underline{71.0} & \textbf{73.0}\\
    Glass & 67.0 & 63.0 & 71.0 & \underline{73.0} & 69.0 & {71.0} & \textbf{74.0}\\
    Motion & 65.0 & 61.0 & 68.0 & \textbf{74.0} & \underline{71.0} & 68.0 & {69.0}\\
    Zoom & 64.0 & 65.0 & 64.0 & 70.0 & 68.0 & \underline{71.0} & \textbf{76.0}\\
    Snow & 70.0 & 65.0 & 62.0 & \underline{73.0} & 68.0 & 69.0 & \textbf{74.0}\\
    Frost & 63.0 & 65.0 & 60.0 & \textbf{69.0} & \underline{66.0} & 65.0 & \underline{66.0}\\
    Fog & \textbf{56.0} & 35.0 & 46.0 & 54.0 & 49.0 & 48.0 & \underline{55.0}\\
    Bright & 72.0 & 71.0 & 69.0 & 75.0 & 74.0 & \underline{77.0} & \textbf{78.0}\\
    Contrast & \underline{39.0} & 22.0 & 34.0 & \textbf{40.0} & 32.0 & 29.0 & 34.0\\
    Elastic & 64.0 & 62.0 & 68.0 & \textbf{71.0} & 65.0 & \textbf{71.0} & \underline{70.0}\\
    Pixel & 68.0 & 70.0 & 68.0 & \underline{74.0} & 69.0 & 71.0 & \textbf{76.0}\\
    JPEG & 67.0 & 66.0 & 68.0 & \underline{72.0} & 70.0 & 69.0 & \textbf{76.0}\\
    \midrule
    \textbf{mAcc} & 64.7 & 60.1 & 63.7 & \underline{69.0} & 65.4 & 66.8 & \textbf{69.8}\\
    \bottomrule
\end{tabular}
\end{adjustbox}
\caption{Comparison of certified accuracy at $r=0.0$ (\%) on CIFAR-10-C of severity 3. {We set the highest and runner-up values bold-faced and underlined, respectively.}}
\label{tab:cifar10c_sev3_clean_rotated}
\end{minipage}
\hspace*{\fill}
\end{table*}

\begin{table*}[ht]
\begin{minipage}{.5\textwidth}
\centering
\begin{adjustbox}{height=1.45in}
\begin{tabular}{l|cccccc|c}
    Type & \rot{Gaussian} & \rot{Stability} & \rot{SmoothAdv} & \rot{MACER} & \rot{Consistency} & \rot{SmoothMix} & \rot{\textbf{CAT-RS (Ours)}} \\ 
    \midrule
    Gaussian & 0.402 & 0.342 & 0.504 & 0.468 & 0.505 & \underline{0.510} & \textbf{0.543}\\
    Shot & 0.417 & 0.352 & 0.500 & 0.473 & 0.503 & \underline{0.507} & \textbf{0.541}\\
    Impulse & 0.376 & 0.308 & 0.490 & 0.442 & 0.489 & \underline{0.494} & \textbf{0.531}\\
    Defocus & 0.360 & 0.320 & 0.474 & 0.432 & 0.477 & \underline{0.484} & \textbf{0.503}\\
    Glass & 0.313 & 0.271 & \underline{0.474} & 0.386 & 0.461 & 0.469 & \textbf{0.499}\\
    Motion & 0.335 & 0.301 & 0.451 & 0.405 & 0.458 & \underline{0.461} & \textbf{0.481}\\
    Zoom & 0.337 & 0.308 & 0.459 & 0.410 & 0.453 & \underline{0.465} & \textbf{0.493}\\
    Snow & 0.311 & 0.308 & \underline{0.414} & 0.360 & 0.399 & 0.369 & \textbf{0.448}\\
    Frost & 0.270 & 0.282 & \underline{0.400} & 0.349 & 0.362 & 0.369 & \textbf{0.405}\\
    Fog & 0.125 & 0.084 & \underline{0.196} & 0.186 & 0.195 & 0.167 & \textbf{0.214}\\
    Bright & 0.363 & 0.369 & 0.486 & 0.446 & \underline{0.492} & 0.473 & \textbf{0.524}\\
    Contrast & 0.071 & 0.082 & \underline{0.140} & 0.107 & 0.122 & 0.112 & \textbf{0.148}\\
    Elastic & 0.309 & 0.263 & 0.438 & 0.385 & \underline{0.446} & 0.440 & \textbf{0.469}\\
    Pixel & 0.389 & 0.345 & 0.498 & 0.460 & 0.496 & \underline{0.509} & \textbf{0.532}\\
    JPEG & 0.412 & 0.352 & \underline{0.503} & 0.465 & 0.500 & 0.501 & \textbf{0.535}\\
    \midrule
    \textbf{mACR} & 0.319 & 0.286 & \underline{0.428} & 0.385 & 0.424 & 0.422 & \textbf{0.458}\\
    \bottomrule
\end{tabular}
\end{adjustbox}
\caption{Comparison of \emph{average certified radius} (ACR) on CIFAR-10-C of severity 4. We set the highest values bold-faced for each row. We set the runner-up values underlined.}
\label{tab:cifar10c_sev4_rotated}
\end{minipage}
\hspace*{\fill}
\begin{minipage}{.45\textwidth}
\centering
\begin{adjustbox}{height=1.45in}
\begin{tabular}{l|cccccc|c}
    Type & \rot{Gaussian} & \rot{Stability} & \rot{SmoothAdv} & \rot{MACER} & \rot{Consistency} & \rot{SmoothMix} & \rot{\textbf{CAT-RS (Ours)}} \\ 
    \midrule
    Gaussian & 71.0 & 64.0 & 68.0 & 72.0 & 70.0 & 72.0 & \textbf{79.0}\\
    Shot & 71.0 & 65.0 & 68.0 & 72.0 & 70.0 & \underline{74.0} & \textbf{77.0}\\
    Impulse & 70.0 & 59.0 & 69.0 & \underline{76.0} & 73.0 & 73.0 & \textbf{77.0}\\
    Defocus & 64.0 & 66.0 & 69.0 & \underline{71.0} & 69.0 & \underline{71.0} & \textbf{73.0}\\
    Glass & 64.0 & 62.0 & 70.0 & 72.0 & 70.0 & \textbf{74.0} & \underline{73.0}\\
    Motion & 66.0 & 61.0 & 69.0 & \underline{70.0} & \underline{70.0} & 69.0 & \textbf{72.0}\\
    Zoom & 65.0 & 63.0 & 64.0 & 69.0 & \underline{70.0} & \underline{70.0} & \textbf{76.0}\\
    Snow & 68.0 & 66.0 & 67.0 & \textbf{71.0} & 64.0 & 68.0 & \underline{69.0}\\
    Frost & \underline{69.0} & 60.0 & 64.0 & 64.0 & 65.0 & \textbf{74.0} & \underline{69.0}\\
    Fog & \underline{42.0} & 26.0 & 40.0 & \textbf{45.0} & 40.0 & \underline{42.0} & \textbf{45.0}\\
    Bright & 70.0 & 72.0 & 69.0 & 72.0 & \underline{76.0} & 73.0 & \textbf{77.0}\\
    Contrast & \underline{25.0} & 19.0 & 22.0 & \textbf{29.0} & 21.0 & 24.0 & 23.0\\
    Elastic & 64.0 & 62.0 & 63.0 & 69.0 & 65.0 & \underline{74.0} & \textbf{77.0}\\
    Pixel & 65.0 & 66.0 & 70.0 &\underline{74.0} & 71.0 & 72.0 & \textbf{76.0}\\
    JPEG & 69.0 & 65.0 & 69.0 & 70.0 & 65.0 & \underline{72.0} & \textbf{78.0}\\
    \midrule
    \textbf{mAcc} & 62.9 & 58.4 & 62.7 & 66.4 & 63.9 & \underline{66.8} & \textbf{69.4}\\
    \bottomrule
\end{tabular}
\end{adjustbox}
\caption{Comparison of certified accuracy at $r=0.0$ (\%) on CIFAR-10-C of severity 4. {We set the highest and runner-up values bold-faced and underlined, respectively.}}
\label{tab:cifar10c_sev4_clean_rotated}
\end{minipage}
\hspace*{\fill}
\end{table*}

\begin{table*}[ht]
\begin{minipage}{.5\textwidth}
\centering
\begin{adjustbox}{height=1.45in}
\begin{tabular}{l|cccccc|c}
    Type & \rot{Gaussian} & \rot{Stability} & \rot{SmoothAdv} & \rot{MACER} & \rot{Consistency} & \rot{SmoothMix} & \rot{\textbf{CAT-RS (Ours)}} \\ 
    \midrule
    Gaussian & 0.408 & 0.335 & 0.501 & 0.467 & 0.500 & \underline{0.511} & \textbf{0.540}\\
    Shot & 0.403 & 0.325 & 0.494 & 0.458 & 0.498 & \underline{0.502} & \textbf{0.532}\\
    Impulse & 0.346 & 0.275 & 0.476 & 0.421 & 0.471 & \underline{0.484} & \textbf{0.505}\\
    Defocus & 0.311 & 0.290 & 0.445 & 0.389 & 0.447 & \underline{0.449} & \textbf{0.471}\\
    Glass & 0.308 & 0.269 & 0.449 & 0.372 & 0.451 & \underline{0.464} & \textbf{0.488}\\
    Motion & 0.321 & 0.286 & 0.438 & 0.382 & 0.445 & \underline{0.446} & \textbf{0.471}\\
    Zoom & 0.316 & 0.296 & \underline{0.449} & 0.391 & 0.437 & 0.446 & \textbf{0.475}\\
    Snow & 0.277 & 0.290 & \underline{0.401} & 0.363 & 0.366 & 0.384 & \textbf{0.420}\\
    Frost & 0.248 & 0.236 & \textbf{0.372} & 0.309 & 0.330 & 0.334 & \underline{0.369}\\
    Fog & 0.078 & 0.046 & 0.086 & \underline{0.110} & \textbf{0.112} & 0.100 & 0.104\\
    Bright & 0.301 & 0.335 & 0.415 & 0.400 & \underline{0.430} & 0.409 & \textbf{0.439}\\
    Contrast & 0.046 & 0.058 & 0.087 & 0.079 & \underline{0.093} & 0.075 & \textbf{0.103}\\
    Elastic & 0.313 & 0.280 & 0.458 & 0.398 & \underline{0.466} & 0.462 & \textbf{0.472}\\
    Pixel & 0.386 & 0.332 & 0.486 & 0.453 & 0.488 & \underline{0.503} & \textbf{0.527}\\
    JPEG & 0.405 & 0.350 & \underline{0.504} & 0.466 & 0.500 & 0.502 & \textbf{0.530}\\
    \midrule
    \textbf{mACR} & 0.298 & 0.267 & 0.404 & 0.364 & 0.402 & \underline{0.405} & \textbf{0.430}\\
    \bottomrule
\end{tabular}
\end{adjustbox}
\caption{Comparison of \emph{average certified radius} (ACR) on CIFAR-10-C of severity 5. We set the highest values bold-faced for each row. We set the runner-up values underlined.}
\label{tab:cifar10c_sev5_rotated}
\end{minipage}
\hspace*{\fill}
\begin{minipage}{.45\textwidth}
\centering
\begin{adjustbox}{height=1.45in}
\begin{tabular}{l|cccccc|c}
    Type & \rot{Gaussian} & \rot{Stability} & \rot{SmoothAdv} & \rot{MACER} & \rot{Consistency} & \rot{SmoothMix} & \rot{\textbf{CAT-RS (Ours)}} \\ 
    \midrule
    Gaussian & 71.0 & 61.0 & 71.0 & \underline{74.0} & 71.0 & 73.0 & \textbf{76.0}\\
    Shot & 68.0 & 62.0 & 67.0 & \underline{71.0} & 69.0 & 70.0 & \textbf{77.0}\\
    Impulse & \underline{72.0} & 57.0 & 68.0 & \underline{72.0} & 66.0 & \textbf{74.0} & \textbf{74.0}\\
    Defocus & 62.0 & 61.0 & 67.0 & 68.0 & 69.0 & \underline{70.0} & \textbf{72.0}\\
    Glass & 63.0 & 59.0 & 67.0 & 67.0 & \underline{70.0} & \textbf{74.0} & \underline{70.0}\\
    Motion & 65.0 & 60.0 & 63.0 & \underline{69.0} & 68.0 & 68.0 & \textbf{70.0}\\
    Zoom & 63.0 & 60.0 & 61.0 & 68.0 & \underline{70.0} & \underline{70.0} & \textbf{75.0}\\
    Snow & 57.0 & 58.0 & 59.0 & 59.0 & \textbf{63.0} & \underline{61.0} & 59.0\\
    Frost & 60.0 & 54.0 & 61.0 & \underline{65.0} & 60.0 & \textbf{66.0} & 61.0\\
    Fog & \underline{31.0} & 13.0 & 17.0 & \textbf{33.0} & 28.0 & 28.0 & 27.0\\
    Bright & 68.0 & \underline{71.0} & 65.0 & 69.0 & \textbf{72.0} & 70.0 & 68.0\\
    Contrast & 18.0 & 15.0 & 12.0 & \textbf{23.0} & 16.0 & 16.0 & \underline{19.0}\\
    Elastic & 64.0 & 64.0 & 65.0 & \underline{70.0} & \textbf{71.0} & 69.0 & 69.0\\
    Pixel & 65.0 & 64.0 & 68.0 & \textbf{74.0} & 70.0 & \underline{71.0} & \textbf{74.0}\\
    JPEG & 67.0 & 66.0 & 68.0 & \underline{70.0} & 67.0 & \underline{70.0} & \textbf{75.0}\\
    \midrule
    \textbf{mAcc} & 59.6 & 55.0 & 58.6 & \underline{63.5} & 62.0 & 63.3 & \textbf{64.4}\\
    \bottomrule
\end{tabular}
\end{adjustbox}
\caption{Comparison of certified accuracy at $r=0.0$ (\%) on CIFAR-10-C of severity 5. {We set the highest and runner-up values bold-faced and underlined, respectively.}}
\label{tab:cifar10c_sev5_clean_rotated}
\end{minipage}
\hspace*{\fill}
\end{table*}

\clearpage
\section{Results on MNIST-C}
\label{ap:mnistc}

We perform the evaluation on MNIST-C \citep{mu2019mnistc}, 15 replicas of MNIST \cite{dataset/mnist}, where each replica consists of a different type of corruption (\eg rotate, shear, spatter, etc.). We evaluate the corruption performance of the smoothed classifiers on the full test dataset of MNIST-C after training the base classifiers with MNIST. In this experiment, we use $\sigma=0.25$. Although the improvement of CAT-RS in MNIST-C is less dramatic than in CIFAR-10-C because confidence information is more important in more complex dataset, CAT-RS still achieves the best mACR among the baselines.\footnote{The dataset is hosted at \url{https://zenodo.org/record/3239543\#.YisCti8RpQJ}.}

\begin{figure*}[ht]%
\centering
\subfigure[][Clean]{%
\label{fig:mnistc_a}%
\includegraphics[height=0.75in]{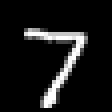}}%
\hspace{5pt}%
\subfigure[][Bright]{%
\label{fig:mnistc_b}%
\includegraphics[height=0.75in]{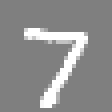}}%
\hspace{5pt}%
\subfigure[][Line]{%
\label{fig:mnistc_c}%
\includegraphics[height=0.75in]{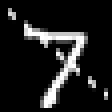}}%
\hspace{5pt}%
\subfigure[][Glass]{%
\label{fig:mnistc_d}%
\includegraphics[height=0.75in]{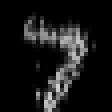}}
\hspace{5pt}%
\subfigure[][Impulse]{%
\label{fig:mnistc_e}%
\includegraphics[height=0.75in]{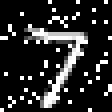}}%
\hspace{5pt}%
\subfigure[][Rotate]{%
\label{fig:mnistc_f}%
\includegraphics[height=0.75in]{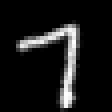}}%
\hspace{5pt}%
\subfigure[][Shear]{%
\label{fig:mnistc_g}%
\includegraphics[height=0.75in]{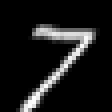}}%
\hspace{5pt}%
\subfigure[][Spatter]{%
\label{fig:mnistc_h}%
\includegraphics[height=0.75in]{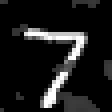}}\\
\subfigure[][Translate]{%
\label{fig:mnistc_i}%
\includegraphics[height=0.75in]{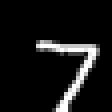}}%
\hspace{5pt}%
\subfigure[][Edges]{%
\label{fig:mnistc_j}%
\includegraphics[height=0.75in]{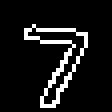}}%
\hspace{5pt}%
\subfigure[][Fog]{%
\label{fig:mnistc_k}%
\includegraphics[height=0.75in]{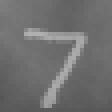}}%
\hspace{5pt}%
\subfigure[][Motion]{%
\label{fig:mnistc_l}%
\includegraphics[height=0.75in]{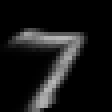}}
\hspace{5pt}%
\subfigure[][Scale]{%
\label{fig:mnistc_m}%
\includegraphics[height=0.75in]{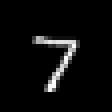}}%
\hspace{5pt}%
\subfigure[][Shot]{%
\label{fig:mnistc_n}%
\includegraphics[height=0.75in]{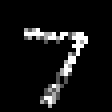}}%
\hspace{5pt}%
\subfigure[][Stripe]{%
\label{fig:mnistc_o}%
\includegraphics[height=0.75in]{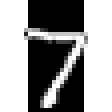}}%
\hspace{5pt}%
\subfigure[][Zigzag]{%
\label{fig:mnistc_p}%
\includegraphics[height=0.75in]{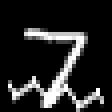}}\\
\caption[Images of MNIST-C.]{Images in MNIST-C test dataset:
\subref{fig:mnistc_a} is a clean test image in MNIST, and the other images are the corresponding corrupted images contained in MNIST-C.}%
\label{fig:mnistc-images}%
\end{figure*}

\begin{table*}[ht]
\begin{minipage}{.5\textwidth}
\centering
\begin{adjustbox}{height=1.45in}
\begin{tabular}{l|cccccc|c}
    Type & \rot{Gaussian} & \rot{Stability} & \rot{SmoothAdv} & \rot{MACER} & \rot{Consistency} & \rot{SmoothMix} & \rot{\textbf{CAT-RS (Ours)}} \\ 
    \midrule
    Bright & 0.540 & \underline{0.599} & 0.320 & \textbf{0.606} & 0.410 & 0.316 & 0.319\\
    Line & 0.856 & 0.865 & 0.906 & 0.867 & 0.885 & \underline{0.901} & \textbf{0.910}\\
    Glass & 0.655 & 0.643 & \underline{0.743} & 0.670 & 0.686 & 0.710 & \textbf{0.758}\\ 
    Impulse & 0.785 & 0.800 & \underline{0.868} & 0.813 & 0.828 & 0.847 & \textbf{0.876}\\
    Rotate & 0.762 & 0.776 & \underline{0.833} & 0.793 & 0.822 & {0.831} & \textbf{0.835} \\
    Shear & 0.850 & 0.857 & \underline{0.900} & 0.869 & 0.891 & 0.899 & \textbf{0.902}\\
    Spatter & 0.841 & 0.844 & \underline{0.895} & 0.860 & 0.880 & 0.892 & \textbf{0.902}\\
    Translate & 0.315 & 0.332 & \underline{0.392} & 0.346 & 0.388 & \textbf{0.449} & 0.366\\
    Edges &0.354 & 0.390 & \underline{0.496} & 0.430 & 0.489 & 0.486 & \textbf{0.519}\\
    Fog & 0.116 & 0.097 & 0.108 & \textbf{0.123} & 0.094 & 0.102 & \underline{0.112}\\
    Motion & 0.626 & 0.610 & \underline{0.704} & 0.627 & 0.675 & \textbf{0.730} & \underline{0.704}\\
    Scale & 0.637 & 0.636 & {0.727} & 0.666 & \underline{0.736} & \textbf{0.766} & 0.714\\
    Shot & 0.836 & 0.835 & \underline{0.902} & 0.856 & 0.886 & 0.894 & \textbf{0.907}\\
    Stripe & 0.532 & 0.590 & 0.678 & 0.700 & \textbf{0.771} & 0.736 & \underline{0.759}\\
    Zigzag & 0.726 & 0.740 & \underline{0.794} & 0.746 & 0.779 & 0.774 & \textbf{0.815}\\
    \midrule 
    \textbf{mACR} & 0.629 & 0.641 & {0.684} & 0.665 & 0.681 & \underline{0.689} &\textbf{0.693}\\
    \bottomrule
\end{tabular}
\end{adjustbox}
\caption{Comparison of \emph{average certified radius} (ACR) on MNIST-C. We set the highest values bold-faced for each row. We set the runner-up values underlined.}
\label{tab:mnistc_rotated}
\end{minipage}
\hspace*{\fill}
\begin{minipage}{.45\textwidth}
\centering
\begin{adjustbox}{height=1.45in}
\begin{tabular}{l|cccccc|c}
    Type & \rot{Gaussian} & \rot{Stability} & \rot{SmoothAdv} & \rot{MACER} & \rot{Consistency} & \rot{SmoothMix} & \rot{\textbf{CAT-RS (Ours)}} \\ 
    \midrule
    Bright & \underline{91.6} & 98.1 & 68.7 & \textbf{97.1} & 82.0 & 63.1 & 64.5\\
    Line & 98.5 & 98.7 & \textbf{99.1} & 98.6 & \underline{98.9} & \textbf{99.1} &\textbf{99.1}\\
    Glass & 96.6 & 96.6 & \textbf{97.3} & \underline{96.8} & 96.7 & 96.6 & \textbf{97.3}\\ 
    Impulse & 97.9 & 98.3 & \textbf{98.9} & 98.5 & \underline{98.7} & \underline{98.7} & \textbf{98.9}\\
    Rotate &92.5 & 93.2 & \underline{94.4} & 93.6 & \underline{94.4} & \textbf{94.7} & 94.1 \\
    Shear & 97.4 & 97.9 & \underline{98.4} & 98.1 & 98.3 & \textbf{98.5} & 98.3\\
    Spatter & 97.9 & 98.1 & \underline{98.8} & 98.3 & \underline{98.8} & \textbf{98.9} & \textbf{98.9}\\
    Translate & 51.7 & 52.8 & 55.6 & 53.4 & \underline{56.6} & \textbf{64.6} & 51.4\\
    Edges & 72.3 & 71.9 & 72.1 & \textbf{75.1} & 73.5 & 72.2 & \underline{73.8} \\
    Fog & 54.7 & \underline{55.8} & 35.2 & \textbf{62.2} & 35.0 & 24.8 & 35.8\\
    Motion &94.7 & 94.8 & 95.9 & 94.9 & \underline{96.2} & \textbf{97.1} & 95.1\\
    Scale & 94.0 & 94.3 & 93.4 & 94.9 & \underline{95.8} & \textbf{96.2} & 91.6\\
    Shot & 98.6 & 98.6 & \underline{99.0} & 98.8 & \textbf{99.1} & \underline{99.0} & \underline{99.0}\\
    Stripe & 76.8 & 81.7 & 88.2 & 89.9 & \textbf{94.0} & \underline{92.5} & 92.0\\
    Zigzag & 90.2 & 91.9 & \underline{93.6} & 91.2 & 92.9 & 93.1 & \textbf{95.2}\\
    \midrule 
    \textbf{mAcc} & 87.0 & \underline{88.2} & 85.9 & \textbf{89.4} & 87.4 & 85.9 & {85.7}\\
    \bottomrule
\end{tabular}
\end{adjustbox}
\caption{Comparison of certified accuracy at $r=0.0$ (\%) on MNIST-C. We set the highest values bold-faced for each row, and the runner-up values underlined.}
\label{tab:mnistc_clean_rotated}
\end{minipage}
\hspace*{\fill}
\end{table*}

\end{document}